\documentclass{article}

\usepackage[preprint,eandd]{neurips}

\usepackage[utf8]{inputenc}
\usepackage[T1]{fontenc}
\usepackage{hyperref}
\usepackage{url}
\usepackage{booktabs}
\usepackage{amsmath,amssymb,amsfonts}
\usepackage{nicefrac}
\usepackage{microtype}
\usepackage{graphicx}
\usepackage{xcolor}
\usepackage{enumitem}
\usepackage[capitalize,noabbrev]{cleveref}

\title{ChaosNetBench: Benchmarking Spatio-Temporal\\
Graph Neural Networks on Chaotic Lattice Dynamics}

\author{
  Henok Tenaw Moges\thanks{Benchmark and Evaluations available at : \href{https://github.com/htmoges/ChaosNetBench}{https://github.com/htmoges/ChaosNetBench}} \\
  Centre for Artificial Intelligence Research (CAIR) \\
  University of Cape Town, Cape Town, South Africa \\
  \texttt{henok.moges@uct.ac.za}
  \And
  Charalampos Skokos \\
  Nonlinear Dynamics and Chaos Group, Department of Mathematics and Applied Mathematics \\
  University of Cape Town, Cape Town, South Africa \\
  \texttt{haris.skokos@uct.ac.za}
  \And
  Deshendran Moodley \\
  Centre for Artificial Intelligence Research (CAIR) \\
  University of Cape Town, Cape Town, South Africa \\
  \texttt{deshen.moodley@uct.ac.za}
}

\begin{document}

\maketitle

\begin{abstract}

Spatio-temporal graph neural networks (STGNNs) are widely used for short-term forecasting in dynamic physical systems such as traffic and weather. However, the prevailing evaluation practice uses real world benchmark data sets in a single domain with a single fixed holdout splits, making it difficult to compare architectures across different dynamical regimes. We introduce ChaosNetBench (CNB), a synthetic benchmark dataset and evaluation framework for studying STGNN performance under controlled multidimensional chaotic dynamics. CNB is built on a lattice of coupled standard maps with independently tunable local chaos ($K$), coupling strength ($\varepsilon$), and system size ($N$), providing known topology and known dynamics across 96 system instances and 9{,}600 trajectories. We introduce chaos indicators, evaluation metrics and a protocol to analyze and compare the capacity of STGNN architectures to deal with different levels of local and global chaos. We illustrate the usage of the framework by analyzing 13 architectures (5 STGNNs and 8 non-graph baselines). The results reveal a regime dependent transition in which non-graph baselines (TCN, N-BEATS, iTransformer) remain competitive when there is low local chaos, while STGNNs (e.g., Graph WaveNet, D2STGNN, STAEformer) are generally more resilient to higher levels of local and global chaos. CNB provides a practical, reusable testbed for systematically comparing and analyzing the capacity of STGNN architectures to handle different levels of local and global chaos. 

\end{abstract}

\section{Introduction}
\label{sec:intro}
Spatio-temporal graph neural networks (STGNNs) have become a dominant approach for short-term forecasting in dynamical physical systems, achieving strong performance across different application domains including traffic and weather~\citep{li2018dcrnn,wu2019graphwavenet,wu2020mtgnn,shao2022d2stgnn,liu2023staeformer, DavidsonM22, gaibie2024predicting}. These models typically combine a temporal module to capture local system dynamics and a graph neural network, based on message passing, to capture inter-system or global dynamics~\citep{cini2025graph}. 

Despite their empirical success, current evaluation practices for STGNNs face two interconnected limitations. First, STGNNs are primarily evaluated on real world benchmark data sets, usually in a single domain such as traffic, which limits the analysis of their generalizability. Reported improvements can vary substantially across application domains, where more recent methods that perform well on traffic benchmarks do not always maintain the same performance in other domains, e.g. the stock market~\citep{pillay2021exploring}. Second, there is limited evaluation for temporal generalization. Most of the widely used datasets, including PEMS-Bay and METR-LA~\citep{li2018diffusion}, and PEMS03, PEMS04, PEMS07 and PEMS08~\citep{song2020spatial} contain traffic data ranging from 2 to 6 months. A standard holdout evaluation typically uses either a 6:2:2 or 7:1:2 split, so trained models are evaluated on at most the last two months of data. While this facilitates comparison of findings across studies, it is unclear whether these models would yield similar performance at other time intervals, which are likely to exhibit different dynamics. At a fundamental level, STGNNs should be assessed on their ability to approximate the dynamics of the underlying physical system.

These limitations motivate the need for a controlled benchmark to analyze the limits of different STGNN architectures on domain-agnostic physical systems. Beyond predictions, models are expected to provide insight into system dynamics, further motivating the need for evaluation settings where behavior can be interpreted relative to a known system structure~\citep{durstewitz2023reconstructing,durstewitz2026position}. Recent work evaluated and compared large scale deep neural networks for predicting single-dimensional chaotic systems~\citep{gilpin2023model}. They showed that transformers, recurrent neural networks, and physics-inspired neural networks achieved strong predictive performance when sufficient training data is available. We extend this perspective to the spatio-temporal setting, to analyze when STGNNs provide advantages over temporal models in multidimensional chaotic systems. 

To this end, we introduce ChaosNetBench (CNB), a controlled benchmark for analyzing STGNN architectures on multi-dimensional chaotic dynamical systems. CNB comprises a synthetic dataset and evaluation framework based on coupled standard maps (CSM), a deterministic, time-invariant lattice system with known ring topology and independently tunable local chaos ($K$), coupling ($\varepsilon$), and system size ($N$)~\citep{chirikov1979,kantz1988arnold,moges2022coupled}. The dataset is generated systematically across values of $K$, $\varepsilon$, and $N$, enabling evaluation across dynamical regimes with varying local instability and spatial coupling strengths. Unlike standard STGNN benchmarks that evaluate a single real world trajectory over time, CNB evaluates multiple trajectories generated from different initial conditions (ICs), allowing performance to be assessed across both independent trajectories and dynamical regimes. Using CNB, we evaluate a diverse set of architectures, from sequential architectures to STGNNs, to analyze when the addition of spatial modules outperforms their temporal counterparts and how robust they remain under increasing levels of local and global chaos.

Three empirical findings emerge from this controlled setting. First, we identify a regime dependent transition governing when STGNNs begin to outperform non-graph baselines, with non-graph models remaining competitive at low coupling relative to local chaos. Second, model comparisons depend on the evaluation view: models with similar averaged accuracy can differ substantially in how reliably they maintain useful predictions at higher levels of chaos. Third, the type of spatial propagation mechanism emerges as a key design factor, with diffusion based mechanisms outperforming state dependent alternatives in highly chaotic regimes.

Beyond these empirical results, the benchmarking framework aims to deepen the analysis of STGNNs to shape better scientific findings when comparing the capacity and limits of different STGNNs architectures for spatio-temporal modelling of physical systems with different levels of complexity. Because the data generating process is fully specified, different evaluation summaries, including averaged rankings, conditional accuracy, and robustness, can be interpreted relative to known dynamical difficulty (\S\ref{sec:framework}).

\textbf{Contributions.}
(1) A controlled benchmarking framework for evaluating STGNNs on coupled chaotic dynamics with known topology and tunable regimes.
(2) A systematic analysis of model behavior across dynamical regimes, highlighting differences between averaged accuracy, conditional accuracy, and robustness.
(3) An evaluation framework that separates complementary views of performance under autoregressive rollout, grounded in known chaos indicators.

\vspace{-1.5mm}
\section{Related Work} \label{sec:related}
\vspace{-1.5mm}

Gilpin's \textsc{dysts} collection~\citep{gilpin2021dysts} provides a standardized library of chaotic dynamical systems for benchmarking forecasting methods across diverse temporal attractors. Building on this, \citet{gilpin2023model} showed that increasing model scale alone does not eliminate the need for domain knowledge, with compact models incorporating structural priors matching larger neural networks on short-term prediction. These works establish that structural assumptions can be beneficial, but they focus on independently evolving systems without explicit spatial coupling.

Prior work has also shown that data-driven models, including reservoir computing approaches, can recover short-term Lyapunov behavior in Spatio-temporal systems~\citep{pathak2018model,Vlachas2021}. However, without controlled topology and tunable coupling, the contribution of spatial modeling remains difficult to isolate.

ChaosBench~\citep{nathaniel2024chaosbench} evaluates weather emulators such as GraphCast~\citep{lam2023learning}, ClimaX~\citep{nguyen2023climax}, and PanguWeather~\citep{bi2023pangu} on ERA5 data for subseasonal to seasonal forecasting. While it provides large scale real world evaluation, the underlying dynamics and topology are not analytically specified, and difficulty must be inferred from proxy variables. CNB addresses a complementary setting by using a controlled synthetic system with known dynamics and orbit-level difficulty labels.

More broadly, synthetic benchmarks such as PDEBench~\citep{takamoto2022pdebench} and XAI-Bench~\citep{liu2021xaibench} leverage known structure to enable interpretable evaluation. Our work follows this paradigm, focusing specifically on coupled chaotic lattice systems and STGNN evaluation under controlled regimes.

STGNNs such as DCRNN~\citep{li2018dcrnn}, Graph WaveNet~\citep{wu2019graphwavenet}, and AGCRN~\citep{bai2020adaptive} establish diffusion based message passing and adaptive graph learning as strong baselines. More recent models introduce increasingly expressive mechanisms: D2STGNN~\citep{shao2022d2stgnn} separates diffusion and local dynamics via learned gating, while STAEformer~\citep{liu2023staeformer} employs state dependent spatial attention. These models are primarily evaluated on traffic and weather benchmarks where the coupling structure and dynamics are unknown, making it difficult to determine whether gains arise from spatial modeling or dataset specific factors. Beyond empirical performance, theoretical analysis of STGNNs is limited. Cini et al.~\cite{cini2023taming} analyze the interplay between local and global dynamics in STGNNs using stochastic processes, AZ-Whiteness testing~\cite{Zambon2022AZwhiteness}, and introduce a synthetic GPVAR benchmark. Our approach is different in that it assumes the underlying physical system is deterministic but may appear random.

A complementary line of work incorporates physical structure directly into learning via Hamiltonian, Lagrangian, neural ODE, and equation discovery approaches~\citep{brunton2016sindy,chen2018neuralode,greydanus2019hamiltonian,cranmer2020lagrangian}. Our objective is orthogonal: rather than enforcing physical structure in the model, we use a controlled dynamical system to evaluate how general purpose forecasting architectures behave under coupled chaotic dynamics with known topology. 

Overall, existing benchmarks either study temporal chaos without spatial coupling or evaluate spatial forecasting without controlled dynamics. As summarized in Table~\ref{tab:benchmark_gap}, no prior benchmark combines spatial coupling, known topology, tunable chaotic regimes, and orbit level difficulty labels. CNB addresses this gap by enabling controlled, identifiable evaluation of STGNNs under chaotic dynamics.

\begin{table}[ht]
\centering
\caption{Benchmark gap. CNB uniquely combines spatial coupling, known topology (analytically specified adjacency), tunable chaotic regimes, orbit level difficulty labels, and a controlled synthetic design enabling identifiable evaluation.}
\label{tab:benchmark_gap}
\footnotesize
\setlength{\tabcolsep}{2pt}
\begin{tabular}{lccccc p{2.6cm}}
\toprule
Benchmark
  & \shortstack{Synthetic\\type}
  & \shortstack{Spatial\\coupling}
  & \shortstack{Known\\topology}
  & \shortstack{Tunable\\chaos}
  & \shortstack{orbit level\\labels}
  & Evaluation target \\
\midrule
Gilpin (2021, 2023)~\citep{gilpin2021dysts,gilpin2023model}
  & $\checkmark$ & $\times$ & $\times$ & $\checkmark$ & $\times$
  & Temporal chaotic forecasting \\
ChaosBench~\citep{nathaniel2024chaosbench}
  & $\times$ & partial$^\dagger$ & $\times$ & $\times$ & proxy$^\dagger$
  & S2S climate forecasting \\
Traffic (METR-LA, PEMS-BAY)
  & $\times$ & partial$^\dagger$ & $\times$ & $\times$ & $\times$
  & Short-term traffic forecasting \\
\textbf{CNB (ours)}
  & $\checkmark$ & $\checkmark$ & $\checkmark$ & $\checkmark$ & $\checkmark$
  & \shortstack[l]{STGNN evaluation\\under controlled chaos} \\
\bottomrule
\end{tabular}
{\footnotesize $^\dagger$Partial: spatial structure present but not analytically specified or controlled. Proxy: difficulty metrics inferred from physical variables, not orbit level ground truth classification.}
\end{table}

\section{The ChaosNetBench Dataset}
\label{sec:framework}

\subsection{Dynamical Chaotic System}
\label{sec:physical_system}
Following the CSM~\citep{chirikov1979,kantz1988arnold,moges2022coupled}, we study $N$ sites arranged on a ring with periodic boundary conditions. Each site $i=1,2,\dots,N$ has a position $q_n^{(i)}$ and momentum $p_n^{(i)}$, updated by

\begin{align} \label{eq:csm}
  p_{n+1}^{(i)} &= p_n^{(i)} + K\sin\!\left(q_n^{(i)}\right)
  - \varepsilon\!\left[\sin\!\left(q_n^{(i+1)} - q_n^{(i)}\right)
  + \sin\!\left(q_n^{(i-1)} - q_n^{(i)}\right)\right], \\
  q_{n+1}^{(i)} &= q_n^{(i)} + p_{n+1}^{(i)} \pmod{2\pi}, \nonumber
\end{align}
with $n$ indicating the map's iteration. The dynamics combine local nonlinear evolution with diffusion-like nearest neighbor coupling on a known, simple ring topology, where each node connects to exactly two neighbors. This makes the CSM a controlled testbed: the adjacency is analytically specified, $K$ and $\varepsilon$ independently modulate local chaos and coupling, and the map defines the physical transition directly (rather than a numerical ODE surrogate). The CSM is an area-preserving map and the discrete-time counterpart of autonomous continuous-time Hamiltonian systems, conserving phase space volume; its orbits can be regular or chaotic depending on $K$ and $\varepsilon$. Orbit-level variability can be systematically characterized using the chaos indicators introduced in \S\ref{sec:eval_framework}.

We define a \emph{coupling-to-chaos ratio} $\rho = \varepsilon/K$, which captures the balance between node-local instability and spatial interaction. When $\varepsilon \ll K$, sites evolve nearly independently; as $\rho$ increases, spatial coupling becomes increasingly influential. Empirically, $\rho$ provides a compact summary of how this balance shifts across regimes. Our maximum ratio $\rho = 0.50$ keeps $\varepsilon \leq K/2$ for all $K$ values, so coupling remains subordinate to the local nonlinear term $K\sin(q)$.\footnote{At $K = 6.5$ with $\rho \geq 0.30$, the system enters a strongly chaotic regime with uniformly short predictability horizons across models.}

\begin{figure}[ht]
  \centering
  \includegraphics[width=0.90\textwidth]{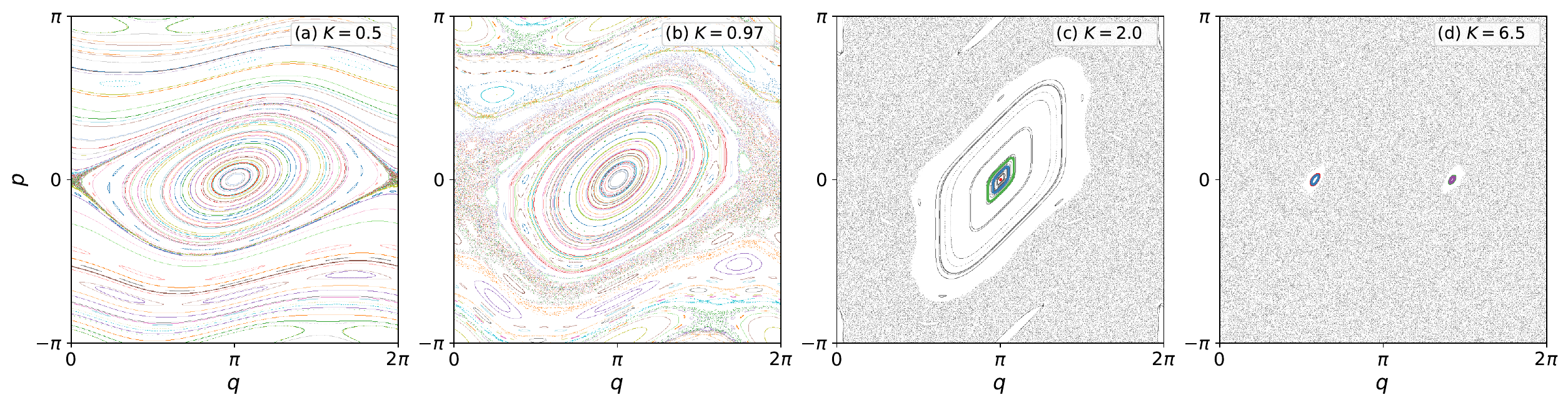}
  \caption{Phase space of the uncoupled ($\varepsilon = 0$) standard map of Eq.~\ref{eq:csm}. Increasing $K$ drives a transition from near-integrable motion ($K=0.5$) to extended chaos ($K=6.5$), motivating the four benchmark regimes.}
  \label{fig:phase_portraits}
\end{figure}

We select four representative values of $K$ spanning distinct uncoupled phase space regimes, from near-integrable motion to extended chaos. Figure~\ref{fig:phase_portraits} shows the phase space of the uncoupled map ($\varepsilon = 0$) across these values: as $K$ increases, invariant structures break down and chaotic regions dominate, defining the regime structure used in the benchmark. This choice motivates the core forecasting question: as local chaos intensifies, how does the role of spatial coupling change?

\subsection{The Dataset}
\label{sec:dataset}
We construct the CNB dataset by sampling the controlled design space defined by the dynamical system in \S\ref{sec:physical_system}. Specifically, we generate \textbf{96 system instances} from four values of $K$, eight values of $\rho$ ($0.05$--$0.50$), and three lattice sizes ($N \in \{8, 16, 32\}$), with 100 ICs per configuration (9{,}600 trajectories total; see Appendix~\ref{app:dataset}, especially Table~\ref{tab:setup}).

As coupling increases, the coupled system statistics shift: both maximum Lyapunov exponent growth and chaos fraction vary systematically with $\rho$ (Table~\ref{tab:regimes}). Chaos indicators (defined in \S\ref{sec:eval_framework}) provide physically grounded, orbit level difficulty labels for each trajectory (Appendix~\ref{app:chaos}). The dataset therefore spans regimes from near-integrable to fully chaotic dynamics across coupling strengths.

We split the dataset by IC, rather than along the time axis. For each system instance, 100 IC trajectories are partitioned into 70 train ICs, 10 validation ICs, and 20 test ICs; windows are generated only after this trajectory level split. This ensures that test trajectories are independent of those seen during training. It differs from standard STGNN benchmark preprocessing, which typically applies temporal 70/10/20 splits to a single multivariate time series~\citep{wu2019graphwavenet}. For a time-invariant deterministic system such as Eq.~\ref{eq:csm}, temporally adjacent segments of the same trajectory remain strongly correlated, particularly under chaotic dynamics; time-based splits can therefore introduce information leakage and overestimate predictive performance. Our IC-based split evaluates generalization across ICs, which is more appropriate for assessing how well models generalize in chaotic systems and is consistent with prior work on learning dynamical systems from multiple trajectories~\citep{gilpin2021dysts,gilpin2023model}.

\begin{figure}[ht]
  \centering
  \includegraphics[width=0.90\textwidth]{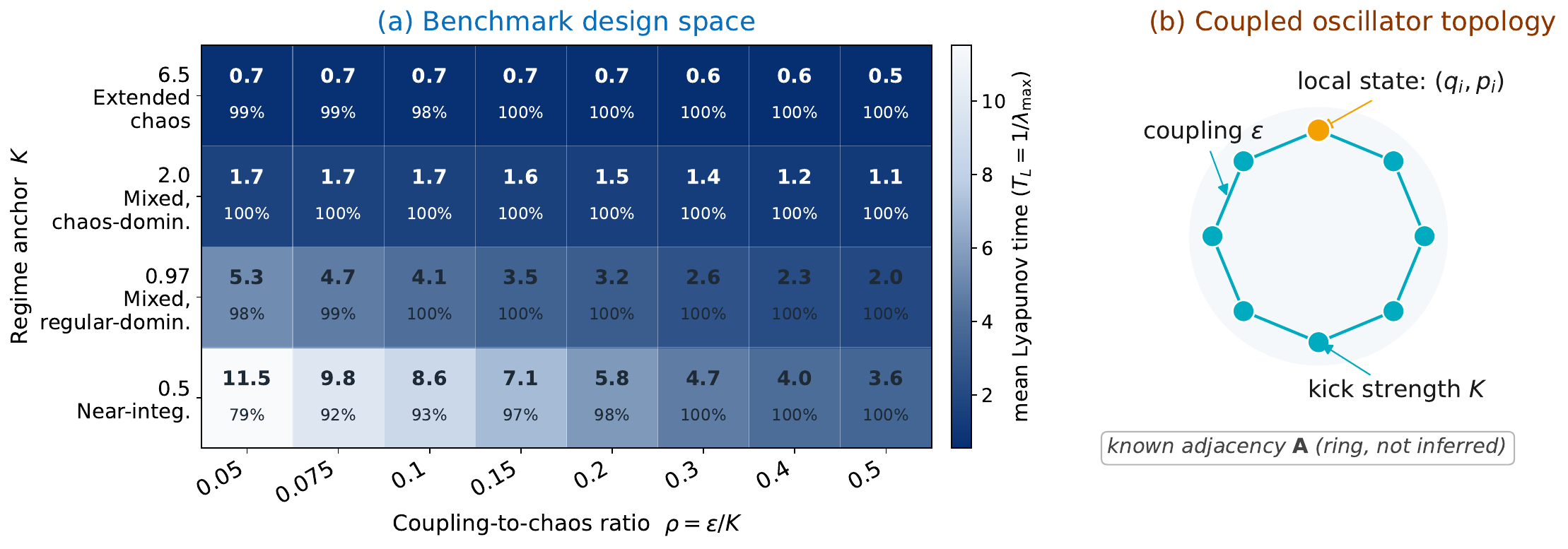}
  \caption{CNB benchmark overview. (a)~96 instance design space ($N=8$ slice shown; the grid repeats for $N\in\{8,16,32\}$), colored by mean Lyapunov time $T_L = 1/\lambda_{\max}$; each cell is labeled with $T_L$ and chaotic fraction. (b)~Ring topology of the CSM with node state $(q_i,p_i)$, coupling $\varepsilon$, local nonlinearity $K$, and analytically specified adjacency $\mathbf{A}$.}
  \label{fig:benchmark_overview_submission}
\end{figure}

Figure~\ref{fig:benchmark_overview_submission} summarizes the benchmark. Panel~(a) maps the 96 instance design space (for $N=8$) with cell color indicating mean Lyapunov time $T_L = 1/\lambda_{\max}$, formally defined in \S\ref{sec:eval_framework}, and labels showing both $T_L$ and chaotic fraction. Panel~(b) shows the ring topology with analytically specified adjacency. This highlights how increasing $K$ and $\rho$ jointly reduce predictability across the benchmark design space.

We summarize regime statistics using the maximal Lyapunov exponent $\lambda_{\max}$ and SALI based orbit classification (defined in \S\ref{sec:eval_framework}). Table~\ref{tab:regimes} shows that increasing $K$ systematically increases the $\lambda_{\max}$ range and drives the system toward fully chaotic behavior, while increasing $\rho$ further amplifies instability across configurations. Higher $K$ regimes exhibit uniformly high chaos fractions and shorter predictability horizons.

\begin{table}[ht]
\centering
\caption{Regime characterization for CNB at $N=8$. The uncoupled column summarizes the standard map at $\varepsilon{=}0$; coupled statistics summarize 100 ICs per configuration over eight $\rho$ values. Extended SALI threshold details and larger $N$ diagnostics are provided in Appendix~\ref{app:chaos}.}
\label{tab:regimes}
\small
\setlength{\tabcolsep}{6pt}
\begin{tabular}{cccc}
\toprule
$K$ & Uncoupled character & \multicolumn{2}{c}{Coupled system (range over $\rho \in [0.05, 0.50]$)} \\
\cmidrule(lr){3-4}
 & & $\bar{\lambda}_{\max}$ range & Chaos fraction (SALI, \S\ref{sec:eval_framework}) \\
\midrule
0.50 & Near-integrable & 0.09--0.28 & 79--100\% \\
0.97 & Mixed, regular dominant & 0.19--0.49 & 98--100\% \\
2.00 & Mixed, chaos dominant & 0.59--0.87 & 100\% \\
6.50 & Extended chaos & 1.40--1.82 & 98--100\% \\
\bottomrule
\end{tabular}
\end{table}

\subsection{Benchmark Task and Evaluation Metrics}
\label{sec:eval_framework}

\textbf{Problem formulation.}
We cast the benchmark as a standard STGNN forecasting problem. Let $\mathcal{G} = (\mathcal{V}, \mathcal{E}, \mathbf{A})$ be a graph with $|\mathcal{V}| = N$ nodes and adjacency $\mathbf{A} \in \{0,1\}^{N \times N}$ encoding the ring topology from Eq.~\ref{eq:csm}. Each node connects to exactly two neighbors, in contrast to real world graphs (e.g., traffic) which typically exhibit one-to-many connectivity. At each step $n$, node $i$ carries a 2-dimensional state $(q_n^{(i)}, p_n^{(i)})$, and the full system state is $\mathbf{X}_n \in \mathbb{R}^{N \times 2}$. Given a context window of $L$ steps, the forecasting task is

\begin{equation} \label{eq:stgnn_task}
  \hat{\mathbf{X}}_{n+1:n+H} = f_\theta\!\left(\mathcal{G},\, \mathbf{X}_{n-L+1:n}\right),
\end{equation}
where $H$ is the forecast horizon.

This interface follows standard STGNN formulations such as DCRNN and Graph WaveNet~\citep{li2018dcrnn,wu2019graphwavenet}, but differs from typical benchmarks in three key aspects: (i)~$\mathbf{A}$ is analytically known and provided with the dataset, rather than inferred from data; (ii)~trajectories are generated by a deterministic, time-invariant map providing exact ground truth evolution; and (iii)~models are evaluated under rollout, where forecast error growth arises from intrinsic trajectory divergence rather than observational noise. Models may use the provided adjacency directly or learn an adaptive one; the benchmark does not require graph recovery as an objective, allowing us to isolate the effect of using spatial structure from topology estimation.

\textbf{Evaluation metrics.}
Standard STGNN benchmarks report Mean Square Error (MSE) and Root Mean Square Error (RMSE) as primary metrics, and CNB reports these for comparability. However, under rollout in chaotic systems, these metrics are dominated by exponential forecast error growth and primarily reflect divergence over time.

We therefore use \textbf{valid prediction time (VPT)} as the primary rollout metric. Following~\citet{pathak2018model}, VPT is defined as the number of rollout steps until the normalized RMSE (NRMSE), normalized by the standard deviation of the true trajectory, first exceeds $1.0$. This corresponds to the point at which predictions lose correlation with the true trajectory and become comparable to the scale of the underlying dynamics. VPT measures how long a model produces informative predictions and aligns evaluation with the finite predictability horizon imposed by chaotic dynamics. For example, at $K=2.0$ ($\bar{\lambda}_{\max} \approx 0.59$), an input window of $L=48$ spans approximately 28 Lyapunov times, leaving limited extrapolation headroom even with substantial context.

\textbf{Validity criterion and two-view evaluation.}
To exclude degenerate forecasts, we define a \textbf{valid rollout} as one where the test MSE remains below $0.95$ over the evaluation horizon. Under unit-variance normalization, a trivial mean predictor yields test MSE~$\approx 1.0$, so this threshold filters predictions that collapse to mean-state or low-variance solutions that do not track the underlying dynamics.

This criterion enables two complementary evaluation perspectives: \textbf{(i) Conditional accuracy:} VPT evaluated only on system instances where both models under comparison produce valid rollouts, compared using a Wilcoxon signed-rank test on matched pairs. \textbf{(ii) Robustness:} the fraction of 96 system instances where a model produces a valid rollout, compared using McNemar's exact test on the 2$\times$2 contingency table.

This separation allows models with similar predictive accuracy to be distinguished by how consistently they produce valid rollouts (see Appendix~\ref{app:impl_protocol} for full protocol). Pairwise comparisons are computed per system instance by comparing rollout performance (e.g., VPT) between models; a ``win'' is assigned to the model with highest performance, with ties distributed equally as fractions.

\textbf{Chaos indicators.}
Each trajectory is characterized by physically grounded indicators of predictability computed from the underlying dynamics, independent of model predictions. The maximal Lyapunov exponent $\lambda_{\max}$~\citep{benettin1980lyapunovb,Skokos2010} quantifies exponential divergence of nearby trajectories; the Lyapunov time $T_L = 1/\lambda_{\max}$ is the characteristic timescale on which the system loses predictability due to sensitive dependence on ICs. SALI~\citep{skokos2001alignment,moges2022coupled} classifies orbits as regular, sticky, or chaotic, providing orbit level difficulty labels in the coupled standard map setting. These indicators enable evaluation outcomes to be interpreted relative to known dynamical difficulty.

\textbf{Evaluation protocol.}
CNB is designed as a plug-in benchmark for STGNN forecasting models. Given the released dataset, evaluation proceeds as follows:
\begin{enumerate}[leftmargin=1.5em,itemsep=0pt,topsep=2pt]
  \item \textbf{Select system instances} from the design space $(K, \rho, N)$ and apply the IC split (70 train, 10 val, 20 test trajectories per instance).
  \item \textbf{Normalize and window the data} by fitting scaling parameters on train trajectories only and generating supervised windows with context $L{=}48$ and horizon $H{=}12$.
  \item \textbf{Train the model} on these windows with fixed hyperparameters across all system instances.
  \item \textbf{Evaluate under rollout} by iteratively feeding model predictions back as input over held-out test trajectories.
  \item \textbf{Apply the validity screen} (test MSE $< 0.95$) to identify valid rollouts.
  \item \textbf{Compute VPT} (NRMSE threshold $= 1.0$) on valid rollouts; report MSE/RMSE for comparability.
  \item \textbf{Report results} under both conditional accuracy (Wilcoxon) and robustness (McNemar).
\end{enumerate}
All parameters governing data generation, splitting, normalization, training, and evaluation are fixed across system instances. Full implementation details are summarized in Appendix~\ref{app:impl_protocol}.


\section{Benchmarking Experiments}
\label{sec:models}

\subsection{Experimental Design}
\label{sec:experimental_design}
We evaluate 13 forecasting models: eight non-graph baselines, four STGNNs, and Oracle GCN (GCN propagation~\citep{kipf2017semi}) as a diagnostic control; see Appendix~\ref{app:models}. The non-graph baselines are DLinear~\citep{zeng2023transformers}, TCN~\citep{lea2017temporal}, N-BEATS~\citep{oreshkin2020nbeats}, PatchTST~\citep{nie2023patchtst}, STID, LSTM~\citep{hochreiter1997lstm}, iTransformer (ITrans)~\citep{liu2023itransformer}, and DSformer~\citep{yu2023dsformer}. The STGNNs are AGCRN~\citep{bai2020adaptive}, Graph WaveNet~\citep{wu2019graphwavenet} (GWNet), D2STGNN~\citep{shao2022d2stgnn}, and STAEformer~\citep{liu2023staeformer}.

The analysis is organized around matched comparisons that isolate how cross-node information is used. TCN vs.\ GWNet tests the effect of adding diffusion based spatial propagation to a strong temporal backbone. PatchTST vs.\ ITrans tests implicit cross-variate attention. GWNet vs.\ STAEformer compares fixed diffusion based propagation with state dependent spatial attention. D2STGNN provides a comparison point for architectures that separate diffusion and local dynamics through a learned gate, and DSformer provides a non-graph attention based contrast within the non-graph group.

All models follow a fixed hyperparameter protocol. Learning rates are tuned once on a single representative configuration ($K{=}0.5$, $\rho{=}0.30$, $N{=}8$) and fixed across the full sweep. A model \emph{completes} a system instance if training finishes without numerical failure, and it produces a \emph{valid rollout} if it also passes the test-MSE threshold ($< 0.95$) defined in \S\ref{sec:eval_framework}. This distinction matters in high chaos regimes, where training may converge but forecasts can still collapse to low-variance or mean-state predictions.

\subsection{Results}
\label{sec:results}
We report three findings that describe how model behavior changes across dynamical regimes, how evaluation summaries influence comparisons, and how different STGNN designs behave under rollout. 

\subsubsection{Coupling–Chaos Crossover: When Spatial Structure Helps}
\label{sec:crossover}
\begin{figure}[ht]
\centering
\includegraphics[width=0.90\textwidth]{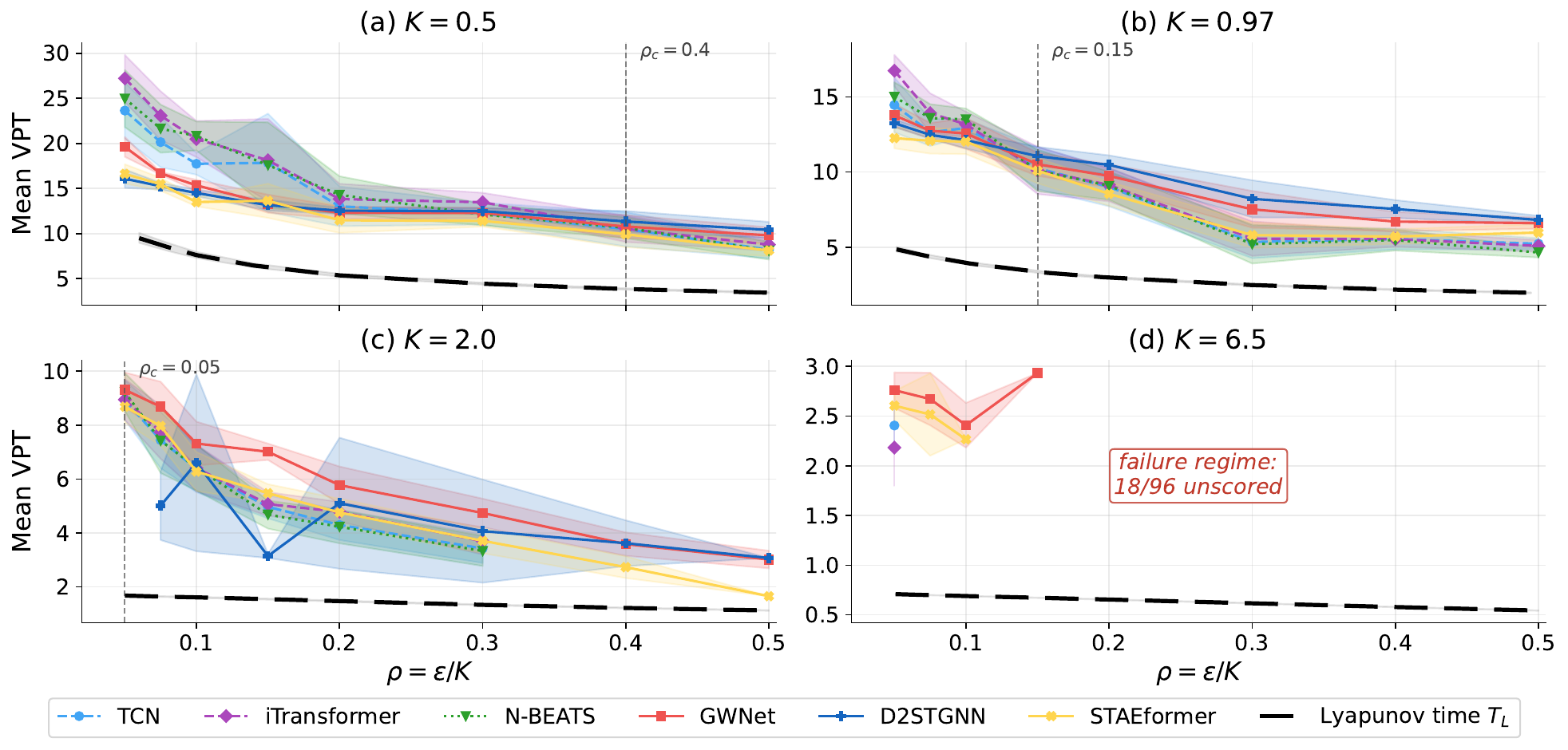}
\caption{Coupling-chaos crossover: mean VPT vs.\ $\rho$ (averaged over
  $N$ ($8, 16, 32$), $\pm 1$ s.e.\ shaded). Non-graph baselines remain competitive at $K{=}0.5$, while STGNNs dominate as $K$ increases. The dashed line marks $T_L = 1/\lambda_{\max}$.}
\label{fig:vpt_rho_crossover}
\end{figure}

Figure~\ref{fig:vpt_rho_crossover} shows the mean VPT for the main architectures across coupling ratio $\rho$ and different chaos regimes $K$, averaged over system sizes $N = 8, 16, 32$ ($\pm 1$ s.e.\ shaded). The results reveal a clear coupling-chaos crossover: non-graph baselines remain competitive in weakly coupled regimes, while STGNNs become more beneficial as coupling increases.

We then summarize the top performing models across the 96 system instances. Using three independent runs per model, we focus on the 78 instances where at least two models produce valid rollouts. D2STGNN (25.0 fractional wins) and GWNet (24.5) appear nearly tied overall, with ITrans trailing (15.0). This ranking masks strong regime dependence (Oracle GCN excluded as a diagnostic control; \S\ref{sec:experimental_design}). The full ranking is shown in Appendix~\ref{fig:overall_ranking_appendix}.

At low local chaos ($K=0.5$), non-graph models dominate: ITrans leads with 10.0 wins, followed by D2STGNN (7.0). At mixed, regular-dominant $K=0.97$, STGNNs become competitive, with D2STGNN leading (14.0 wins). At mixed, chaos-dominant $K=2.0$, GWNet becomes dominant (17/23 scored instances). At extended chaos $K=6.5$, only seven scored instances remain, with GWNet winning 5.5.  

Table~\ref{tab:rho_c} summarizes this transition. The dominant model group shifts from non-graph baselines at low chaos to STGNNs at higher chaos, while the crossover threshold $\rho_c(K)$ moves toward smaller values as $K$ increases. Figure~\ref{fig:vpt_rho_crossover} further shows mean VPT increasing with coupling for STGNNs and approaching the Lyapunov time $T_L$.

\begin{table}[t]
\centering
\caption{Estimated crossover threshold in the Oracle-excluded primary ranking (diagnostic control removed). The regime leader shifts from non-graph models at low chaos to STGNNs at higher chaos. The $K=6.5$ row is a failure analysis regime.}
\label{tab:rho_c}
\small
\setlength{\tabcolsep}{7pt}
\begin{tabular}{cccc}
\toprule
$K$ & Estimated $\rho_c(K)$ & Regime leader & Dominant model group \\
\midrule
0.5 & $\approx 0.40$ & ITrans (10.0 wins) & Non-graph baselines \\
0.97 & $\approx 0.15$ & D2STGNN (14.0 wins) & STGNNs \\
2.0 & $\lesssim 0.05$ & GWNet (17.0 wins) & STGNNs \\
6.5 & $\to 0$ (scored only) & GWNet (5.5 wins) & STGNNs  \\
\bottomrule
\end{tabular}
\end{table}

This pattern reflects the balance between local instability and spatial interaction. When $\varepsilon \ll K$, node-local dynamics dominate and cross-node information provides limited predictive value. As $\rho$ increases, coupling becomes comparable to local divergence, and graph structure becomes informative. This effect disappears at $K \geq 2.0$, where STGNNs consistently outperform non-graph baselines across all $\rho$ (Appendix Figure~\ref{fig:size_crossover_submission}).

\subsubsection{Conditional accuracy and robustness}
\label{sec:robustness_divergence}

\begin{figure}[ht]
\centering
\includegraphics[width=0.90\textwidth]{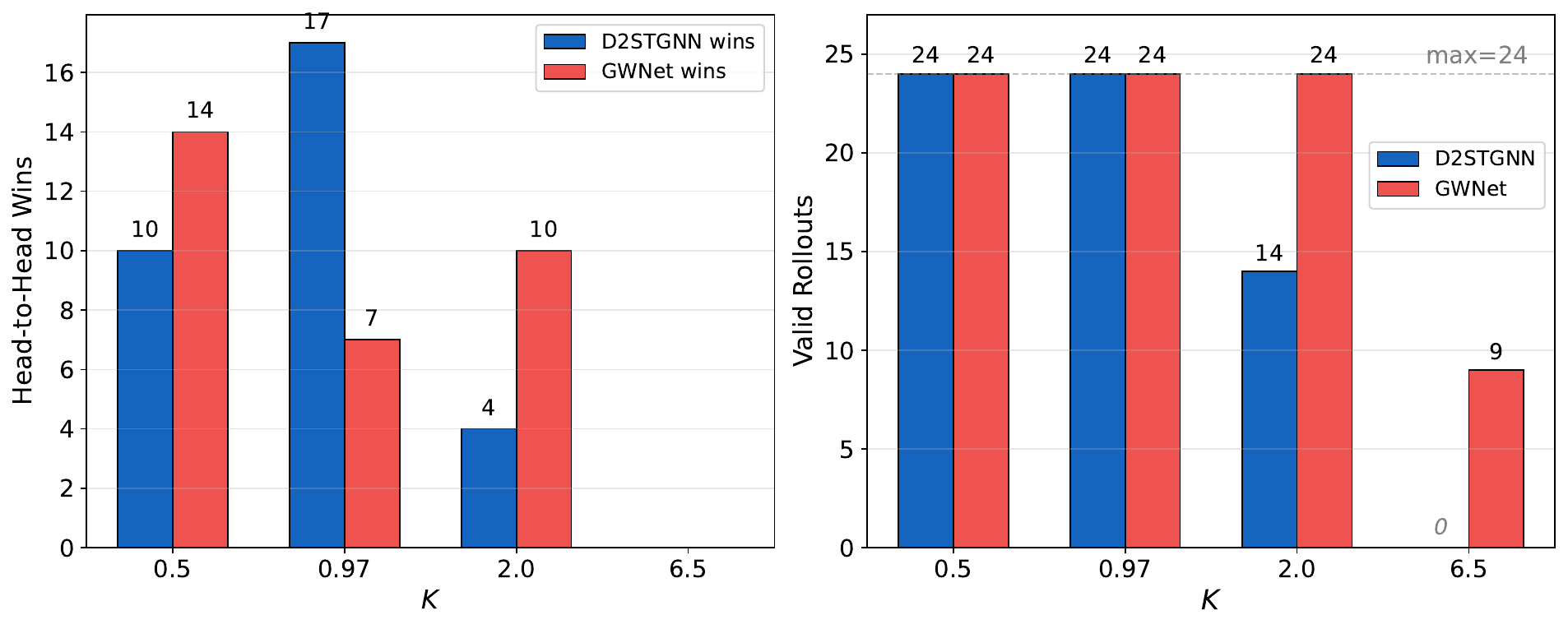}
\caption{D2STGNN vs.\ GWNet (Left) head-to-head VPT wins on the 62 instances where both models produce valid rollouts; (Right) valid rollouts by chaos regime. GWNet matches D2STGNN in conditional accuracy but remains valid on 19 additional instances}
\label{fig:d2stgnn_gwnet_submission}
\end{figure}

Figure~\ref{fig:d2stgnn_gwnet_submission} compares D2STGNN and GWNet under conditional accuracy and robustness evaluation. Across the 62 instances where both models produce valid rollouts, D2STGNN and GWNet are statistically indistinguishable (31--31 head-to-head, Wilcoxon $p = 0.15$). The advantage is regime dependent: D2STGNN leads at $K{=}0.97$ (17--7), while GWNet leads at $K{=}0.5$ (14--10) and $K{=}2.0$ (10--4).

In contrast, robustness differs substantially between the two models. D2STGNN produces valid rollouts on 62/96 system instances (24/24/14/0 at $K{=}0.5/0.97/2.0/6.5$), whereas GWNet does so on 81/96 (24/24/24/9). D2STGNN fails on 19 instances where GWNet succeeds, with no instances of the reverse. A McNemar test (62 both, 0 D2-only, 19 GW-only, 15 neither) yields $p = 3.8 \times 10^{-6}$, indicating a statistically significant robustness gap.

These differences become more pronounced as chaos increases. D2STGNN shows an advantage at $K{=}0.97$ in conditional accuracy, but its robustness drops sharply in more chaotic regimes. At $K{=}2.0$, it trails 4--10 head-to-head against GWNet while showing higher seed variance ($\sigma$ up to 7.07 vs.\ 0.12--0.32 for GWNet), suggesting greater instability. More broadly, CNB shows that conditional accuracy and robustness can diverge substantially across regimes.

\subsubsection{Stability of Spatial Information Propagation Under Chaos}
\label{sec:ablations}

Table~\ref{tab:trainability_submission} summarizes robustness in the high-chaos regimes. STAEformer completes all 96 runs and produces valid rollouts on 77/96 instances, yet wins only 1.5 of 78 scored comparisons. Across the 77 instances where both models produce valid rollouts, GWNet dominates 71--5 (1 tie; $p = 4.5 \times 10^{-12}$), indicating a substantial gap in predictive stability.

\begin{table}[ht]
\centering
\caption{Robustness under increasing chaos. GWNet remains the most stable model, while STAEformer often produces valid rollouts without competitive VPT.}
\label{tab:trainability_submission}
\small
\setlength{\tabcolsep}{4pt}
\begin{tabular*}{\linewidth}{@{\extracolsep{\fill}}lccp{0.48\linewidth}}
\toprule
Model & $K=2.0$ & $K=6.5$ & Interpretation \\
\midrule
GWNet & 24/24 & 9/24 & Most robust across the high-chaos regimes \\
STAEformer & 22/24 & 7/24 & Often valid, but rarely competitive on VPT \\
AGCRN & 18/24 & 0/24 & Learned diffusion degrades in high chaos \\
LSTM & 18/24 & 0/24 & Temporal baseline collapses in high chaos \\
DSformer & 14/24 & 0/24 & Cross-variate attention without an explicit graph \\
D2STGNN & 14/24 & 0/24 & Strong conditional accuracy with reduced robustness \\
\bottomrule
\end{tabular*}
\end{table}

The contrast becomes most visible in the extended chaos regime. While several models continue to produce valid rollouts at $K=2.0$, only GWNet and STAEformer maintain nonzero robustness at $K=6.5$. However, GWNet retains substantially stronger predictive performance across scored comparisons.

These results suggest that propagation mechanisms conditioned on intermediate states are more sensitive to error amplification as chaos increases. Additional matched comparisons are reported in Appendix~\ref{app:results}.

\section{Discussion}
\label{sec:discussion}
ChaosNetBench provides a controlled benchmark built from a synthetic dataset and evaluation framework. 
Across the 96 system instance sweep, the CNB framework reveals a regime dependent crossover in which non-graph baselines remain competitive at low coupling relative to local chaos, while STGNNs become more beneficial as local chaos and coupling increase. It also shows that STGNN architectures with similar conditional accuracy can differ sharply in robustness to higher levels of local and global chaos, and that diffusion based propagation mechanisms remain more stable than state dependent propagation in high chaos regimes.

These comparisons are interpretable because the CNB dataset is generated by a deterministic, time-invariant system with known topology and evaluated on hold-out trajectories under rollout. As a complement to real world STGNN datasets, the framework separates predictive quality from the ability to maintain valid rollouts in a setting where dynamics, architecture, and evaluation can be studied systematically.

\section{Limitations}
\label{sec:limitations}
The evaluation methodology, separating conditional accuracy from robustness, does not depend on the specific dynamics of the CSM and applies broadly. The regime-specific findings, however, are system-dependent: the crossover thresholds $\rho_c(K)$, relative model rankings, and observed failure modes reflect the CSM's local ring coupling and homogeneous structure.

Three aspects of the current setup limit generalization. First, the CSM is a homogeneous lattice with local diffusion-like coupling. Results support a claim about this class of systems, rather than a universal statement about STGNN behavior on arbitrary graph topologies. Whether the observed diffusion advantage persists under heterogeneous graphs (e.g., scale-free networks) remains an open question. Second, the robustness gap observed for D2STGNN is not fully explained. Elevated seed variance suggests optimization fragility, but the attribution to the gating mechanism is observational rather than supported by systematic ablation. Targeted hyperparameter tuning across regimes may alter this behavior and remains an open direction. Third, the dataset assumes noise-free observations. While this isolates the effect of dynamical instability, it does not capture measurement noise usually present in real world systems. Incorporating controlled noise models (e.g. Gaussian noise) is straightforward within this framework and would extend evaluation to partial observability. 

\section{Conclusion}
\label{sec:conclusion}
By combining a deterministic, time-invariant system with known topology, IC-based splits, and rollout evaluation, ChaosNetBench (CNB) provides a reusable framework to complement real world benchmark datasets for analyzing the capacity of different STGNN architectures to handle different levels of local and global chaos. We illustrate its use on 13 architectures across 96 chaotic coupled standard map system instances. The results show that the spatial module in STGNNs increases their capacity over purely temporal approaches for modeling multi-dimensional physical systems. Considering both conditional accuracy and robustness further shows that the performance improvement is not uniform across different STGNN architectures, with diffusion-based propagation mechanisms appearing to be more stable under extended chaos.

\section*{Acknowledgments}
H.~T.~M. and D.~M. acknowledge the financial support by the National Research Foundation (NRF) of South Africa (Grant No. 151217). The authors also thank the the University of Cape Town High Performance Computing (HPC) facility for providing the computational resources used in this work.

\bibliographystyle{unsrtnat}
\bibliography{bibliography}

@article{nathaniel2024chaosbench,
  title={Chaosbench: A multi-channel, physics-based benchmark for subseasonal-to-seasonal climate prediction},
  author={Nathaniel, Juan and Qu, Yongquan and Nguyen, Tung and Yu, Sungduk and Busecke, Julius and Grover, Aditya and Gentine, Pierre},
  journal={Advances in Neural Information Processing Systems},
  volume={37},
  pages={43715--43729},
  year={2024},
}

@article{takamoto2022pdebench,
  title={Pdebench: An extensive benchmark for scientific machine learning},
  author={Takamoto, Makoto and Praditia, Timothy and Leiteritz, Raphael and MacKinlay, Daniel and Alesiani, Francesco and Pfl{\"u}ger, Dirk and Niepert, Mathias},
  journal={Advances in neural information processing systems},
  volume={35},
  pages={1596--1611},
  year={2022}
}

@inproceedings{liu2021xaibench,
  title={Synthetic benchmarks for scientific research in explainable machine learning},
  author={Liu, Yang and Khandagale, Sujay and White, Colin and Neiswanger, Willie},
  booktitle={Advances in Neural Information Processing Systems Datasets Track},
  year={2021}
}

@article{kantz1988arnold,
  title={Internal {Arnold} diffusion and chaos thresholds in coupled symplectic maps},
  author={Kantz, Holger and Grassberger, Peter},
  journal={Journal of Physics A: Mathematical and General},
  volume={21},
  pages={L127--L133},
  year={1988}
  }

@article{moges2022coupled,
  title={Anomalous diffusion in single and coupled standard maps with extensive chaotic phase spaces},
  author={Moges, Henok T. and Manos, Thanos and Skokos, Charalampos},
  journal={Physica D: Nonlinear Phenomena},
  volume={431},
  pages={133120},
  year={2022}
}

@article{chirikov1979,
  author  = {Chirikov, Boris V.},
  title   = {A universal instability of many-dimensional oscillator systems},
  journal = {Physics Reports},
  volume  = {52},
  pages   = {263--379},
  year    = {1979}
}

@article{benettin1980lyapunovb,
  title={Lyapunov characteristic exponents for smooth dynamical systems and for {H}amiltonian systems; a method for computing all of them. {P}art 2: Numerical application},
  author={Benettin, Giancarlo and Galgani, Luigi and Giorgilli, Antonio and Strelcyn, Jean-Marie},
  journal={Meccanica},
  volume={15},
  pages={21--30},
  year={1980}
}

@article{Skokos2010,
   title = {The {L}yapunov characteristic exponents and their computation},
   author = {Charalampos Skokos},
   journal = {Lecture Notes in Physics},
   pages = {63--135},
   volume = {790},
   year = {2010},
}

@article{skokos2001alignment,
  title={Alignment indices: a new, simple method for determining the ordered or chaotic nature of orbits},
  author={Skokos, Charalampos},
  journal={Journal of Physics A: Mathematical and General},
  volume={34},
  pages={10029--10043},
  year={2001}
}

@article{gilpin2021dysts,
  title={Chaos as an interpretable benchmark for forecasting and data-driven modelling},
  author={Gilpin, William},
  journal={Advances in Neural Information Processing Systems},
  volume={34},
  year={2021}
}

@article{gilpin2023model,
  title={Model scale versus domain knowledge in statistical forecasting of chaotic systems},
  author={Gilpin, William},
  journal={Physical Review Research},
   volume={5},
  pages={043252},
  year={2023}
}

@article{pathak2018model,
  title={Model-free prediction of large spatiotemporally chaotic systems from data: A reservoir computing approach},
  author={Pathak, Jaideep and Hunt, Brian and Girvan, Michelle and Lu, Zhixin and Ott, Edward},
  journal={Physical Review Letters},
  volume={120},
  pages={024102},
  year={2018}
}

@article{Vlachas2021,
  title={Backpropagation algorithms and reservoir computing in recurrent neural networks for the forecasting of complex spatiotemporal dynamics},
  author={Vlachas, Pantelis R. and Pathak, Jaideep and Hunt, Brian R. and Sapsis, Themistoklis P. and Girvan, Michelle and Ott, Edward and Koumoutsakos, Petros},
  journal={Neural Networks},
  volume={126},
  pages={191--217},
  year={2020}
}

@inproceedings{kipf2017semi,
  title={Semi-supervised classification with graph convolutional networks},
  author={Kipf, Thomas N. and Welling, Max},
  booktitle = {International Conference on Learning Representations (ICLR)},
  year={2017}
}

@inproceedings{wu2019graphwavenet,
  title={Graph WaveNet for Deep Spatial-Temporal Graph Modeling},
  author={Wu, Zonghan and Pan, Shirui and Long, Guodong and Jiang, Jing and Zhang, Chengqi},
  booktitle={Proceedings of the Twenty-Eighth International Joint Conference on Artificial Intelligence (IJCAI)},
  pages={1907-1913},
  year={2019}
}

@article{greydanus2019hamiltonian,
  title={Hamiltonian Neural Networks},
  author={Greydanus, Samuel and Dzamba, Misko and Yosinski, Jason},
  journal={Advances in Neural Information Processing Systems},
  volume={32},
  year={2019}
}

@article{chen2018neuralode,
  title={Neural ordinary differential equations},
  author={Chen, Ricky T. Q. and Rubanova, Yulia and Bettencourt, Jesse and Duvenaud, David K.},
  journal={Advances in neural information processing systems},
  volume={31},
  year={2018}
}

@article{shao2022d2stgnn,
  title={Decoupled dynamic spatial-temporal graph neural network for traffic forecasting},
  author={Shao, Zezhi and Zhang, Zhao and Wang, Fei and Wei, Wei and Xu, Yongjun},
  journal={Proceedings of the VLDB Endowment},
  volume={15},
  pages={2733--2746},
  year={2022}
}

@inproceedings{yu2023dsformer,
  title={DDsformer: A double sampling transformer for multivariate time series long-term prediction},
  author={Yu, Chengqing and Wang, Fei and Shao, Zezhi and Sun, Tao and Wu, Lin and Xu, Yongjun},
  booktitle={Proceedings of the 32nd ACM International Conference on Information and Knowledge Management (CIKM)},
  pages={2898--2908},
  year={2023}
  }

@inproceedings{liu2023staeformer,
  title={Spatio-Temporal Adaptive Embedding Makes Vanilla Transformer {SOTA} for Traffic Forecasting},
  author={Liu, Hangchen and Dong, Zheng and Jiang, Renhe and Deng, Jiewen and Deng, Jinliang and Chen, Quanjun and Song, Xuan},
  booktitle={Proceedings of the 32nd ACM International Conference on Information and Knowledge Management (CIKM)},
  pages={4125--4129},
  year={2023}
}

@inproceedings{wu2020mtgnn,
  title={Connecting the dots: Multivariate time series forecasting with graph neural networks},
  author={Wu, Zonghan and Pan, Shirui and Long, Guodong and Jiang, Jing and Chang, Xiaojun and Zhang, Chengqi},
  booktitle={Proceedings of the 26th ACM SIGKDD International Conference on Knowledge Discovery \& Data Mining},
  pages={753--763},
  year={2020}
}

@inproceedings{lea2017temporal,
  title={Temporal convolutional networks for action segmentation and detection},
  author={Lea, Colin and Flynn, Michael D. and Vidal, Rene and Reiter, Austin and Hager, Gregory D.},
  booktitle={proceedings of the IEEE Conference on Computer Vision and Pattern Recognition},
  pages={156--165},
  year={2017}
}

@inproceedings{li2018dcrnn,
  title={Diffusion Convolutional Recurrent Neural Network: Data-Driven Traffic Forecasting},
  author={Li, Yaguang and Yu, Rose and Shahabi, Cyrus and Liu, Yan},
  booktitle={International Conference on Learning Representations (ICLR)},
  year={2018}

}

@article{brunton2016sindy,
  title={Discovering governing equations from data by sparse identification of nonlinear dynamical systems},
  author={Brunton, Steven L. and Proctor, Joshua L. and Kutz, J. Nathan},
  journal={Proceedings of the National Academy of Sciences},
  volume={113},
  pages={3932--3937},
  year={2016}
}

@inproceedings{cranmer2020lagrangian,
  title={Lagrangian Neural Networks},
  author={Cranmer, Miles and Greydanus, Sam and Hoyer, Stephan and Battaglia, Peter and Spergel, David and Ho, Shirley},
  booktitle = {ICLR 2020 Workshop on Deep Differential Equations},
  year={2020}
}

@article{zeng2023transformers,
  title={Are transformers effective for time series forecasting?},
  author={Zeng, Ailing and Chen, Muxi and Zhang, Lei and Xu, Qiang},
  journal={Proceedings of the AAAI Conference on Artificial Intelligence},
  volume={37},
  pages={11121-11128},
  year={2023}
}

@article{bai2020adaptive,
  title={Adaptive Graph Convolutional Recurrent Network for Traffic Forecasting},
  author={Bai, Lei and Yao, Lina and Li, Can and Wang, Xianzhi and Wang, Can},
  journal={Advances in Neural Information Processing Systems (NeurIPS)},
  volume={33},
  pages={17804--17815},
  year={2020}
}

@inproceedings{akhtar2024croissant,
  title={Croissant: A Metadata Format for {ML}-Ready Datasets},
  author={Akhtar, Mubashara and Benjelloun, Omar and Conforti, Costanza and Gijsbers, Pieter and Gonzalez, Joan and Kuchnik, Michael and Lhoest, Quentin and Marcenac, Pierre and Maskey, Manil and Mattson, Peter and Montoya, Luis Oala and Raut, Amit and Shinde, Swapnil and Simperl, Elena and Thomas, Goeffry and Tykhonov, Slava and Vanschoren, Joaquin and Vogt, Jos and Wu, Carole-Jean},
  booktitle={KDD},
  year={2024}
}

@inproceedings{shao2022spatial,
  title={Spatial-temporal identity: A simple yet effective baseline for multivariate time series forecasting},
  author={Shao, Zezhi and Zhang, Zhao and Wang, Fei and Wei, Wei and Xu, Yongjun},
  booktitle={Proceedings of the 31st ACM international conference on information \& knowledge management},
  pages={4454--4458},
  year={2022}
}

@inproceedings{nie2023patchtst,
  title={A Time Series is Worth 64 Words: Long-term Forecasting with Transformers},
  author={Nie, Yuqi and Nguyen, Nam H. and Sinthong, Phanwadee and Kalagnanam, Jayant},
  booktitle={International Conference on Learning Representations (ICLR)},
  year={2023}
}

@inproceedings{liu2023itransformer,
  title={iTransformer: Inverted Transformers Are Effective for Time Series Forecasting},
  author={Liu, Yong and Hu, Tengge and Zhang, Haoran and Wu, Haixu and Wang, Shiyu and Ma, Lintao and Long, Mingsheng},
  booktitle={International Conference on Learning Representations (ICLR)},
  year={2024}
}

@inproceedings{DavidsonM22,
title = {ST-GNNs for Weather Prediction in South Africa},
author = {Mikhail Davidson and Deshendran Moodley},
year = {2022},
booktitle = {Artificial Intelligence Research - Third Southern African Conference,
SACAIR 2022, Stellenbosch, South Africa, December 5-9, 2022, Proceedings},
volume = {1734},
pages = {93–107},
}

@article{oreshkin2020nbeats,
  title={{N-BEATS}: Neural Basis Expansion Analysis for Interpretable Time Series Forecasting},
  author={Oreshkin, Boris N. and Carpov, Dmitri and Chapados, Nicolas and Bengio, Yoshua},
  journal={International Conference on Learning Representations},
  year={2020},
}

@article{hochreiter1997lstm,
  title={Long Short-Term Memory},
  author={Hochreiter, Sepp and Schmidhuber, J{\"u}rgen},
  journal={Neural Computation},
  volume={9},
  pages={1735--1780},
  year={1997},
}

@article{lam2023learning,
  title={Learning skillful medium-range global weather forecasting},
  author={Lam, Remi and Sanchez-Gonzalez, Alvaro and Willson, Matthew and Wirnsberger, Peter and Fortunato, Meire and Alet, Ferran and Ravuri, Suman and Ewalds, Timo and Eaton-Rosen, Zach and Hu, Weihua and others},
  journal={Science},
  volume={382},
  pages={1416--1421},
  year={2023},
}

@inproceedings{nguyen2023climax,
  title={ClimaX: a25 foundation model for weather and climate},
  author={Nguyen, Tung and Brandstetter, Johannes and Kapoor, Ashish and Gupta, Jayesh K and Grover, Aditya},
  booktitle={Proceedings of the 40th International Conference on Machine Learning},
  pages={25904--25938},
  year={2023}
}

@article{bi2023pangu,
  title   = {Accurate medium-range global weather forecasting with 3D neural networks},
  author  = {Bi, Kaifeng and Xie, Lingxi and Zhang, Hengheng and Chen, Xin and Gu, Xiaotao and Tian, Qi},
  journal = {Nature},
  volume  = {619},
  pages   = {533--538},
  year    = {2023}
}

@article{durstewitz2026position,
  title={Position: Why a Dynamical Systems Perspective is Needed to Advance Time Series Modeling},
  author={Durstewitz, Daniel and Hemmer, Christoph J{\"u}rgen and Hess, Florian and Doll, Charlotte Ricarda and Eisenmann, Lukas},
  journal={arXiv preprint arXiv:2602.16864},
  year={2026}
}

@article{durstewitz2023reconstructing,
  title={Reconstructing computational system dynamics from neural data with recurrent neural networks},
  author={Durstewitz, Daniel and Koppe, Georgia and Thurm, Max Ingo},
  journal={Nature Reviews Neuroscience},
  volume={24},
  number={11},
  pages={693--710},
  year={2023},
  publisher={Nature Publishing Group UK London}
}

@article{cini2025graph,
  title={Graph deep learning for time series forecasting},
  author={Cini, Andrea and Marisca, Ivan and Zambon, Daniele and Alippi, Cesare},
  journal={ACM Computing Surveys},
  volume={57},
  pages={1--34},
  year={2025},
}

@inproceedings{gaibie2024predicting,
  title={Predicting and Discovering Weather Patterns in South Africa Using Spatial-Temporal Graph Neural Networks},
  author={Gaibie, Adeeb and Amir, Hamza and Nandutu, Irene and Moodley, Deshendran},
  booktitle={Southern African Conference for Artificial Intelligence Research},
  pages={144-160},
  year={2024},
}

@inproceedings{pillay2021exploring,
  title={Exploring graph neural networks for stock market prediction on the JSE},
  author={Pillay, Kialan and Moodley, Deshendran},
  booktitle={Southern African Conference for Artificial Intelligence Research},
  pages={95--110},
  year={2021},
  organization={Springer}
}

@inproceedings{Zambon2022AZwhiteness,
  title={Az-whiteness test: a test for uncorrelated noise on spatio-temporal graphs},
  author={Zambon, Daniele and Alippi, Cesare},
  booktitle={Advances in Neural Information Processing Systems 35},
  pages={11975--11986},
  year={2022},
}

@article{cini2023taming,
  title={Taming local effects in graph-based spatiotemporal forecasting},
  author={Cini, Andrea and Marisca, Ivan and Zambon, Daniele and Alippi, Cesare},
  journal={Advances in Neural Information Processing Systems},
  volume={36},
  pages={55375--55393},
  year={2023}
}

@inproceedings{li2018diffusion,
  title={Diffusion Convolutional Recurrent Neural Network: Data-Driven Traffic Forecasting},
  author={Li, Yaguang and Yu, Rose and Shahabi, Cyrus and Liu, Yan},
  booktitle={International Conference on Learning Representations (ICLR)},
  year={2018}
}

@inproceedings{song2020spatial,
  title={Spatial-temporal synchronous graph convolutional networks: A new framework for spatial-temporal network data forecasting},
  author={Song, Chao and Lin, Youfang and Guo, Shengnan and Wan, Huaiyu},
  booktitle={Proceedings of the AAAI conference on artificial intelligence},
  volume={34},
  pages={914--921},
  year={2020}
}

\newpage
\appendix

\section{Dataset Details}
\label{app:dataset}

The full parameter grid comprises $4\,K \times 8\,\rho \times 3\,N = 96$ system instances, each with 100 ICs (9{,}600 trajectories in total). Each trajectory is stored as a $(T, 2N)$ array where $T = 10{,}000$ steps and columns are ordered $(q_1, \ldots, q_N, p_1, \ldots, p_N)$. Metadata attributes include $K$, $\varepsilon$, $N$, IC index, random seed (for IC reproducibility), computed $\lambda_{\max}$ and the SALI-based orbit classification. The complete dataset occupies 27.3\,GB (compressed HDF5 format). Croissant metadata records and download instructions are provided in the dataset repository.

\begin{table}[ht]
\centering
\caption{Experimental setup.}
\label{tab:setup}
\small
\setlength{\tabcolsep}{4pt}
\begin{tabular}{lp{0.62\linewidth}}
\toprule
Component & Setting \\
\midrule
System & CSM (Eq.~\ref{eq:csm}) on an $N$-site ring \\
Parameter grid & $K \in \{0.5,\, 0.97,\, 2.0,\, 6.5\}$, $N \in \{8,\, 16,\, 32\}$ \\
 & $\rho = \varepsilon / K \in \{0.05,\, 0.075,\, 0.10,\, 0.15,\, 0.20,\, 0.30,\, 0.40,\, 0.50\}$ \\
State & Raw $(q_i,p_i), i=1,\ldots,N$ coordinates \\
System instances & 96 total \\
Trajectories & 100 ICs per configuration \\
Split & 70/10/20 (by IC) \\
Primary metric & Valid prediction time (VPT), autoregressive rollout \\
Orbit diagnostics & SALI and finite time $\lambda_{\max}$ \\
Valid rollout threshold & Test MSE $< 0.95$ (degenerate rollout screen) \\
Hyperparameters & Selected once on a representative configuration ($K{=}0.5$, $\rho{=}0.30$, $N{=}8$) and then fixed \\
\bottomrule
\end{tabular}
\end{table}

\paragraph{Trajectory generation.}
For each configuration, we simulate 100 independent trajectories from randomly sampled ICs, each of total length $T = 11{,}000$ steps ($1{,}000$ transient $+$ $10{,}000$ recorded). ICs are sampled uniformly: positions $q_0^{(i)} \in [0, 2\pi)$ and momenta $p_0^{(i)} \in [-\pi, \pi)$. The transient is discarded; the remaining $10{,}000$ steps constitute the recorded trajectory.

\paragraph{Data split.}
We split the dataset by the ICs, not by time (70/10/20 for train/val/test), ensuring test trajectories are from entirely independent ICs and preventing temporal data leakage.

\paragraph{Sliding window.}
All models take $L=48$ input steps and forecast the next $H=12$. At $K=2.0$ ($\bar{\lambda}_{\max} \approx 0.59$), $L=48$ corresponds to $\approx 28$ Lyapunov times. Training uses stride~12 (non-overlapping prediction windows) to reduce pseudoreplication.

\paragraph{Autoregressive rollout.}
Each test trajectory is initialized with a ground truth context window ($L=48$). The model predictions are iteratively fed back in blocks of $H=12$ steps until the sequence ends. VPT is the first step where the NRMSE threshold of 1.0 is crossed. A valid rollout threshold (test MSE $< 0.95$) filters uninformative degenerate mean-collapse.

The benchmark overview is given in Figure~\ref{fig:benchmark_overview_submission} (\S\ref{sec:framework}); orbit-difficulty diagnostics are detailed in Appendix~\ref{app:chaos}.

\section{Benchmark Protocol and Implementation Details}
\label{app:impl_protocol}
Table~\ref{tab:impl_protocol} summarizes the key parameters governing data generation, splitting, normalization, training, and evaluation. These settings are fixed across all system instances.

Three design choices distinguish this protocol from standard STGNN benchmark practice and form part of the benchmark design.

\begin{itemize}[leftmargin=1.5em,itemsep=2pt,topsep=2pt]

  \item \textbf{IC-based split, not temporal.}  
  The dataset is split by IC: 70 train / 10 validation / 20 test ICs are sampled independently from the full set of 100. Windows are generated only \emph{after} this partition, ensuring that test windows do not overlap with training data. Standard STGNN benchmarks instead apply temporal splits to a single multivariate series (e.g.~\citep{wu2019graphwavenet,wu2020mtgnn}). For deterministic chaotic systems, temporally adjacent segments remain strongly correlated; temporal splits can therefore inflate apparent generalization. The IC-based protocol evaluates generalization across trajectories and dynamical states.

  \item \textbf{Rollout-based evaluation with VPT.}  
  Models are trained on sliding supervised windows (standard STGNN interface) but evaluated under rollout: ground truth context $L{=}48$ is provided once, after which predictions are recursively fed back in blocks of $H{=}12$ until the trajectory ends. Valid prediction time (VPT, NRMSE threshold $= 1.0$) is the primary metric, reflecting the finite predictability horizon imposed by chaotic divergence. This differs from pointwise horizon evaluation (e.g., MSE/RMSE at fixed $H$) used in standard STGNN benchmarks.

  \item \textbf{Separation of conditional accuracy and robustness.}  
  The validity screen (test MSE $< 0.95$) partitions instances into valid and degenerate rollouts. Conditional accuracy (VPT, Wilcoxon) is computed only on instances where both models produce valid rollouts, while robustness (McNemar) is computed over all 96 instances. This separation exposes failure modes that are not visible in averaged accuracy summaries.

\end{itemize}

\begin{table}[ht]
\centering
\caption{Implementation protocol used in the benchmark experiments. Values follow the benchmark configuration and evaluation pipeline; the reported sweep uses 50 training epochs.}
\label{tab:impl_protocol}
\small
\setlength{\tabcolsep}{5pt}
\begin{tabular}{lll}
\toprule
Component & Value & Description \\
\midrule
\multicolumn{3}{l}{\textit{Dataset generation}} \\
ICs per system instance & 100 & trajectories per system instance \\
Recorded steps per IC & 10{,}000 & after transient removal \\
Discarded transient & 1{,}000 & initial steps removed \\
\midrule
\multicolumn{3}{l}{\textit{IC-based split}} \\
Train ICs & 70 & independent trajectories \\
Val ICs & 10 & held-out for model selection \\
Test ICs & 20 & unseen trajectories \\
Normalization & per-node $z$-score & fit on train ICs only \\
\midrule
\multicolumn{3}{l}{\textit{Supervised window training}} \\
Input context ($L$) & 48 & past steps \\
Forecast horizon ($H$) & 12 & predicted steps \\
Train stride & 12 & non-overlapping windows \\
Val/Test stride & 1 & dense evaluation \\
Epochs & 50 & fixed across models \\
Early stopping & 10 & patience \\
\midrule
\multicolumn{3}{l}{\textit{Rollout evaluation}} \\
Initial context & $L=48$ & ground truth seed \\
Step size & $H=12$ & iterative prediction \\
VPT threshold & 1.0 & NRMSE cutoff \\
Validity threshold & MSE $< 0.95$ & valid rollout criterion \\
\bottomrule
\end{tabular}
\end{table}

\section{Chaos Indicators}
\label{app:chaos}

To quantify orbit stability in the coupled system, we classify trajectories using the SALI~\citep{skokos2001alignment}, computed from each trajectory's IC. In the benchmark dataset, SALI screening is performed over a fixed $n=1000$ tangent-map horizon, with early termination once $\mathrm{SALI} < 10^{-8}$ (see e.g., \citep{moges2022coupled}). This yields a three-tier classification: chaotic ($\mathrm{SALI}<10^{-8}$), sticky ($10^{-8} \leq \mathrm{SALI} \leq 10^{-4}$), and regular ($\mathrm{SALI}>10^{-4}$).

The maximal Lyapunov exponent $\lambda_{\max}$ is computed with the Benettin deviation vector renormalization procedure~\citep{benettin1980lyapunovb} on the full $10{,}000$ step post-transient trajectories. Importantly, the SALI-based labels are determined entirely within the fixed screening horizon, while longer iterations are used only for diagnostic validation.

\begin{figure}[h]
  \centering
  \includegraphics[width=0.97\textwidth]{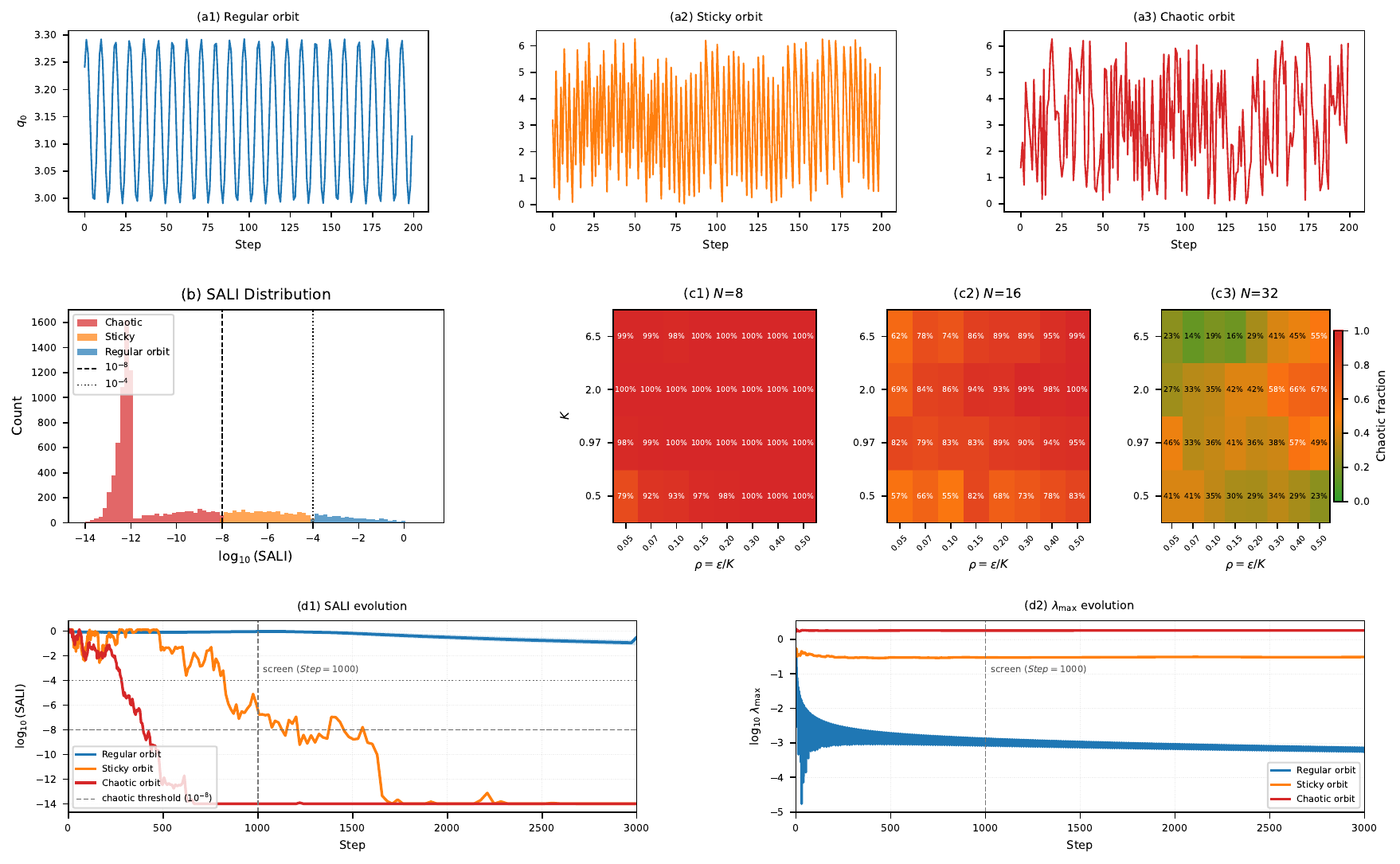}
  \caption{Extended dataset diagnostics: (a)~representative $q_0(t)$ trajectories for the three orbit types: (a1)~regular orbit from the \emph{uncoupled} 2D standard map ($\varepsilon{=}0$, $K{=}0.5$, $q_0{=}\pi{+}0.1$, $p_0{=}0.1$); (a2)~sticky orbit ($K{=}0.97$, $\rho{=}0.15$, $N{=}32$); (a3)~chaotic orbit ($K{=}6.5$, $\rho{=}0.50$, $N{=}8$); (b)~SALI distribution with classification thresholds (chaotic: $\mathrm{SALI}{<}10^{-8}$, regular: $\mathrm{SALI}{>}10^{-4}$); (c)~chaos fraction over $(K,\rho)$ for $N\in\{8,16,32\}$ (c1: $N{=}8$; c2: $N{=}16$; c3: $N{=}32$); (d)~SALI (d1) and running $\lambda_{\max}$ (d2) evolution to $n{=}3000$ steps for the same three exemplars as (a): the chaotic orbit shows rapid SALI collapse and $\lambda_{\max}$ converging to a positive value; the sticky orbit shows intermediate behavior with delayed SALI decrease and saturation to smaller $\lambda_{\max}$; the regular orbit maintains near-constant SALI and decaying $\lambda_{\max}\to 0$. Vertical dashed line marks the $n{=}1000$ screening horizon.}
  \label{fig:dataset_diagnostics_appendix}
\end{figure}

Figure~\ref{fig:dataset_diagnostics_appendix} summarizes the diagnostic structure of the dataset from qualitative regimes to quantitative validation. Panel (a) shows representative $q_0(t)$ trajectories for the three orbit types: regular (a1), sticky (a2), and chaotic (a3), demonstrating the qualitative behavioral differences targeted by the SALI classification. Panel (b) shows the full $\log_{10}(\mathrm{SALI})$ distribution across all screened orbits with classification thresholds overlaid, confirming that the three regimes are well separated. Panels (c1--c3) report the chaos fraction across the $(K,\rho)$ grid for each system size $N\in\{8,16,32\}$, confirming consistent regime ordering across scale.

Panel (d) validates the screening procedure: SALI$(n)$ (d1) and running $\lambda_{\max}(n)$ (d2) are shown up to $n=3000$ steps for the same three representative orbits. All three classes are clearly separated well before the $n=1000$ screening horizon (dashed line), confirming that the fixed-horizon screen is sufficient for classification.

\section{Model Architectures and Hyperparameters}
\label{app:models}

All thirteen models share the frozen hyperparameter protocol described in \S\ref{sec:models}. Architecture details, layer configurations, completion rates, and the per-model learning rates (selected during pilot tuning) are summarized below. Representative parameter counts are listed in the Fairness paragraph below.

\begin{table}[ht]
\centering
\caption{Shared hyperparameter protocol. For each, model learning rates are tuned once on a representative system instance ($K{=}0.5$, $\rho{=}0.30$, $N{=}8$) and fixed for the full sweep.}
\label{tab:hp_protocol}
\small
\setlength{\tabcolsep}{5pt}
\begin{tabular}{lr|lr}
\toprule
HP & Value & HP & Value \\
\midrule
Batch size & 32 & Base LR & $1\times10^{-3}$ \\
Epochs & 50 & LR (LSTM, TCN, D2STGNN) & $3\times10^{-3}$ \\
Scheduler & MultiStepLR & LR (DLinear, N-BEATS, AGCRN) & $3\times10^{-4}$ \\
Milestones & $\{20, 35, 45\}$ & Early stopping & patience=10 \\
$\gamma$ & 0.5 & Weight decay & 0.0 \\
Grad clip & 5.0 & Train stride & 12 \\
Input ($L$) & 48 & Horizon ($H$) & 12 \\
\bottomrule
\end{tabular}
\end{table}

\begin{table}[h]
\centering
\caption{Model suite with spatial processing type, three-seed completion rates, and median per-configuration training time. $^\ddagger$Median wall-clock training time per configuration at $N{=}8$ (hex HPC, NVIDIA A100-80\,GB); includes all epochs to early-stopping.}
\label{tab:model_suite_app}
\footnotesize
\setlength{\tabcolsep}{3pt}
\begin{tabular}{llcrp{3.2cm}}
\toprule
Model & Spatial Processing & 3-seed & \multicolumn{1}{c}{Train$^\ddagger$} & Role \\
 & & & \multicolumn{1}{c}{(min)} & \\
\midrule
DLinear~\citep{zeng2023transformers}     & None (temporal only)     & 96/96 &  10 & Linear baseline \\
TCN~\citep{lea2017temporal}                   & None (temporal only)     & 96/96 & 201 & Negative control (no graph) \\
N-BEATS~\citep{oreshkin2020nbeats}       & None (temporal only)     & 96/96 &  84 & MLP baseline \\
PatchTST~\citep{nie2023patchtst}         & None (temporal only)     & 96/96 &  91 & Self-attention ablation \\
STID~\citep{shao2022spatial}             & Node ID embedding        & 96/96 &  48 & Identity control \\
LSTM~\citep{hochreiter1997lstm}          & Cross-variate, implicit  & 96/96 &  40 & Recurrent diagnostic \\
ITrans~\citep{liu2023itransformer}       & Cross-variate, implicit  & 96/96 &  41 & Cross-variate attention \\
DSformer~\citep{yu2023dsformer}          & Cross-variate, implicit  & 96/96 &  60 & Transformer contrast \\
Oracle GCN~\citep{kipf2017semi}          & Fixed, oracle topology   & 96/96 &  30 & Diagnostic (fixed true adjacency) \\
AGCRN~\citep{bai2020adaptive}            & Learned diffusion        & 96/96 & 326 & Adaptive GCN \\
GWNet~\citep{wu2019graphwavenet}         & Learned diffusion        & 96/96 & 156 & Diffusion convolution \\
D2STGNN~\citep{shao2022d2stgnn}         & Learned diffusion        & 96/96 & 120 & Diffusion/inherent split \\
STAEformer~\citep{liu2023staeformer}     & Learned attention        & 96/96 & 176 & Spatial attention \\
\bottomrule
\end{tabular}
\end{table}

\paragraph{Oracle GCN (diagnostic control).}
Oracle GCN is a lightweight graph convolutional model that uses the analytically specified ring adjacency with fixed symmetric normalization. The architecture consists of (i) a linear temporal embedding, (ii) stacked graph convolution layers with fixed adjacency, and (iii) an MLP output head. This design isolates the effect of correct topology under a simple fixed diffusion mechanism, serving as a diagnostic control for separating topology from propagation and learning effects. It is not intended as a competitive baseline but as a controlled reference for interpreting the role of topology.

\paragraph{Fairness and comparison protocol.}
All models follow their standard published architectures; implementation details are consistent with prior work. We do not enforce parameter count matching across model families. Each family's minimal viable configuration yields inherently different scales (DLinear~$\sim\!1$K, TCN~$\sim\!36$K, N-BEATS~$\sim\!366$K, PatchTST~$\sim\!77$K, STID~$\sim\!117$K, LSTM~$\sim\!58$K, ITrans~$\sim\!71$K, DSformer~$\sim\!20$K, Oracle~GCN~$\sim\!5$K, AGCRN~$\sim\!189$K, GWNet~$\sim\!13$K, D2STGNN~$\sim\!77$K, STAEformer~$\sim\!142$K). Rather than inflating small models or constraining large ones, we follow standard practice of using the smallest reasonable configuration per family and report exact counts.

\paragraph{Training cost and robustness.}
Training cost varies by more than 30-fold across the suite (Table~\ref{tab:model_suite_app}), yet cost does not predict robustness. AGCRN, the most expensive primary model at a median of 326\,min per configuration, achieves only 66/96 valid rollouts at $N{=}8$, lower than all other learned-spatial models, while GWNet reaches 81/96 at 156\,min per configuration. Among temporal baselines, TCN (201\,min) requires the most time: its dilated convolutional stack runs close to the 50-epoch cap on convergent low-$K$ regimes, in contrast to DLinear (10\,min) and LSTM (40\,min). These comparisons underscore that the benchmark's discriminative signal does not require large per-run budgets; architectural choice, not scale, determines robustness.

\section{Statistical Analysis}
\label{app:statistics}

All main comparisons use three seed averages and explicitly separate \emph{conditional accuracy} from \emph{robustness}. Conditional accuracy is evaluated on instances where both models pass the valid rollout threshold (test MSE ${<}\,0.95$), using a two-sided Wilcoxon signed-rank test. Robustness is evaluated over all 96 system instances using a McNemar exact test on the paired valid-rollout table.

Table~\ref{tab:dual_stats} reports the five pre-registered comparisons under both perspectives. This dual view is informative: the same pair of models can be indistinguishable in conditional accuracy yet differ sharply in robustness. The table shows both views together, highlighting where conditional accuracy and robustness lead to different conclusions.

\begin{table}[h]
\centering
\caption{Dual-view statistical evaluation of five pre-registered comparisons.
Conditional accuracy is measured on instances where both models produce valid rollouts (Wilcoxon signed-rank), while robustness is measured on all 96 system
instances (McNemar exact test). The two views capture complementary aspects of
performance and may diverge.}
\label{tab:dual_stats}
\small
\setlength{\tabcolsep}{3pt}
\resizebox{\linewidth}{!}{%
\begin{tabular}{llcccccc}
\toprule
 & & \multicolumn{3}{c}{Conditional accuracy} & \multicolumn{3}{c}{Robustness} \\
\cmidrule(lr){3-5} \cmidrule(lr){6-8}
Comparison & Interpretation & Pairs & Winner & $p$ & $A$-only & $B$-only & McNemar $p$ \\
\midrule
TCN ($A$) vs.\ GWNet ($B$)
& Diffusion ablation
& 69 & GWNet 50--19 & $3.3\times10^{-2}$
& 0 & 12 & $4.9\times10^{-4}$ \\

PatchTST ($A$) vs.\ ITrans ($B$)
& Cross-variate attention
& 66 & ITrans 52--13 (1) & $1.6\times10^{-7}$
& 1 & 2 & $1.0$ \\

GWNet ($A$) vs.\ STAEformer ($B$)
& Fixed vs.\ state dependent propagation
& 77 & GWNet 71--5 (1) & $4.5\times10^{-12}$
& 4 & 0 & $1.3\times10^{-1}$ \\

GWNet ($A$) vs.\ Oracle ($B$)
& Learned vs.\ true graph
& 66 & GWNet 54--12 & $9.1\times10^{-8}$
& 15 & 0 & $6.1\times10^{-5}$ \\

D2STGNN ($A$) vs.\ GWNet ($B$)
& Fidelity vs.\ robustness
& 62 & 31--31 & $1.5\times10^{-1}$
& 0 & 19 & $3.8\times10^{-6}$ \\
\bottomrule
\end{tabular}
}
\end{table}

Across the five comparisons, three conditional-accuracy results remain significant after Bonferroni correction ($\alpha = 0.05/5 = 0.01$). The TCN vs.\ GWNet comparison is significant at $\alpha = 0.05$ but does not survive correction, consistent with its effect being primarily regime dependent (\S\ref{sec:crossover}) and more clearly expressed in robustness (McNemar $p = 4.9 \times 10^{-4}$). In contrast, D2STGNN vs.\ GWNet is non-significant ($p = 0.15$), indicating statistical parity in conditional accuracy, while still exhibiting a pronounced robustness gap (McNemar $p = 3.8 \times 10^{-6}$).

Inter-seed variability is low for most models (median $\sigma \approx 0.22$--$0.30$ steps), supporting stability of the reported rankings. D2STGNN shows the highest variability ($\bar{\sigma} = 0.68$), consistent with the optimization fragility observed in the main analysis, while DLinear is the most stable ($\bar{\sigma} = 0.10$).

\section{Extended Results and Failure-Mode Analysis}
\label{app:results}

This appendix decomposes the main findings into regime wise behavior and exposes mechanism-specific failure modes not visible in overall summaries.

\begin{figure}[h]
  \centering
  \includegraphics[width=0.97\textwidth]{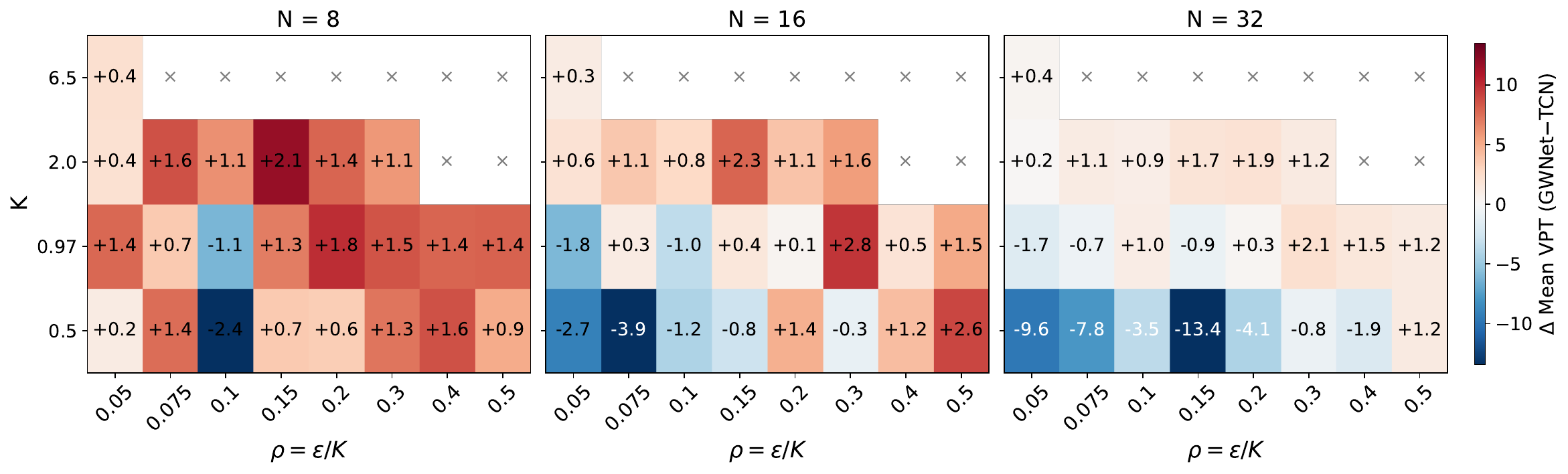}
  \caption{Diffusion-convolution contribution (GWNet minus TCN VPT).
  Blue: explicit diffusion adds little or is slightly harmful; red: diffusion
  is decisively useful.}
  \label{fig:tcn_gwnet_submission}
\end{figure}

\begin{figure}[h]
  \centering
  \includegraphics[width=0.97\textwidth]{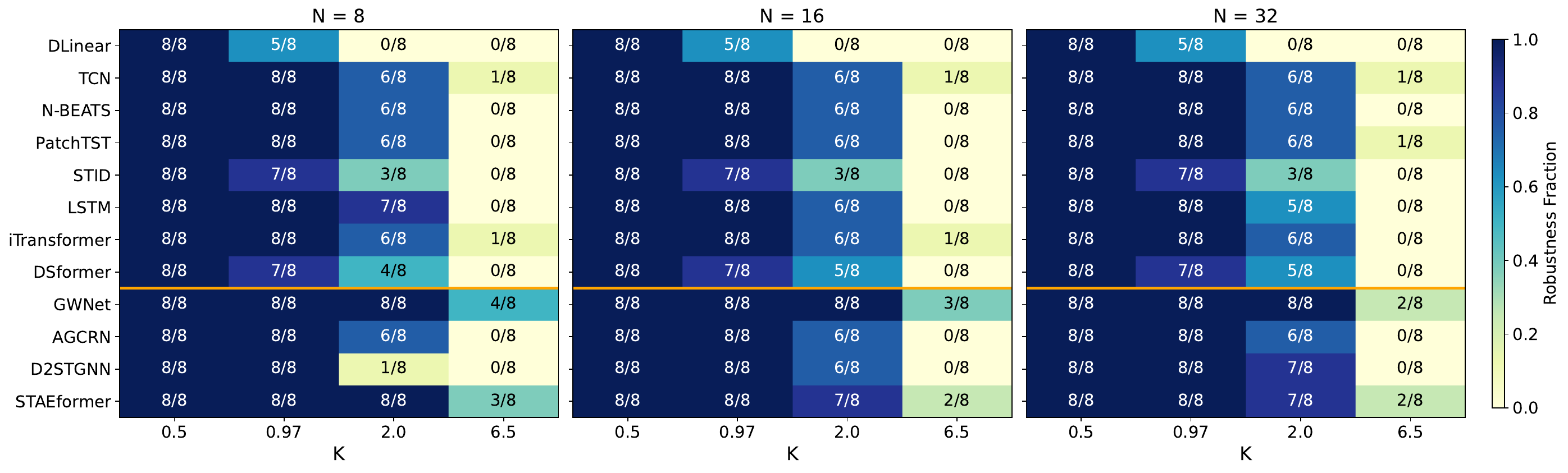}
  \caption{Robustness landscape across chaos levels and system sizes. GWNet
  produces valid rollouts deepest into the high-chaos regime, while D2STGNN and
  DSformer expose mechanism-specific failure modes.}
  \label{fig:trainability_submission}
\end{figure}

This appendix provides extended regime wise breakdowns and pre-registered statistical tests from the three seed confirmation sweep, complementing the main findings on the coupling-chaos crossover and the separation between accuracy and robustness. 

\begin{figure}[h]
  \centering
  \includegraphics[width=0.97\textwidth]{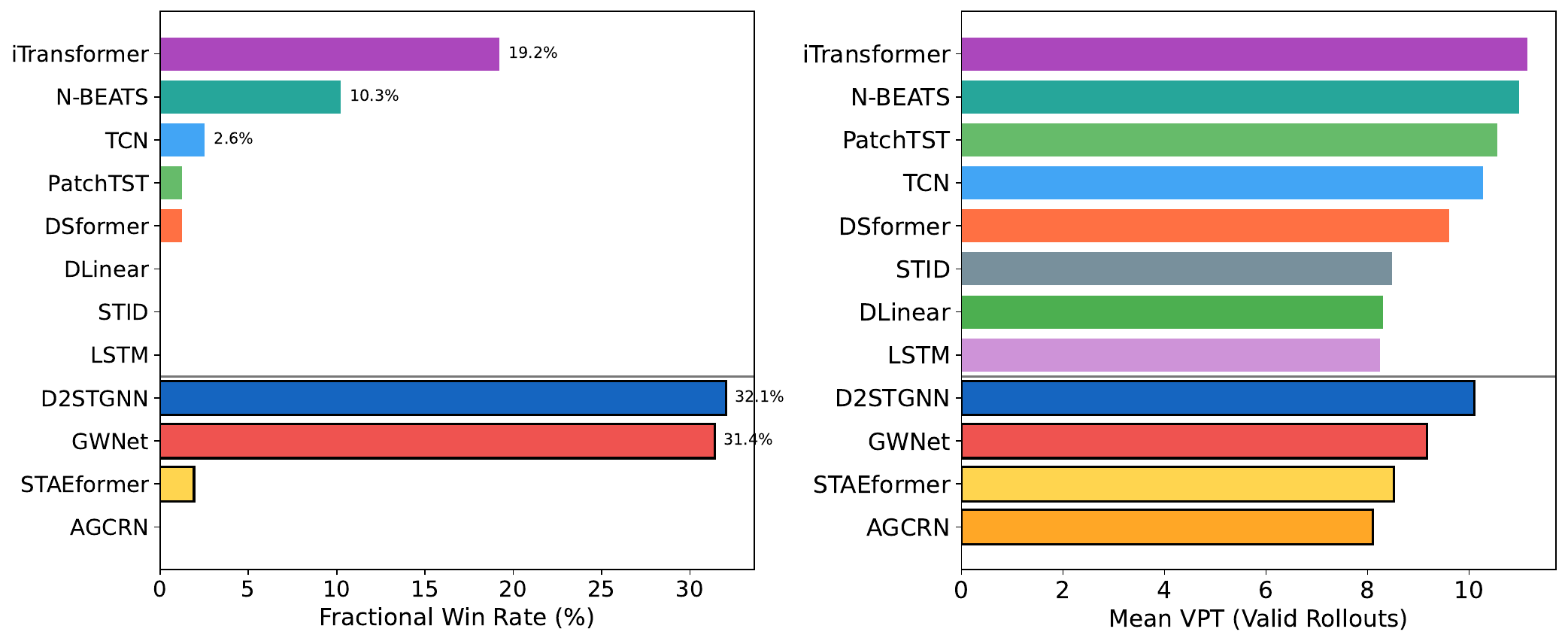}
  \caption{Overall ranking across twelve forecasting models (three-seed means). Left: fractional wins over 78 scored system instances (ties split evenly; Oracle~GCN excluded). A horizontal separator marks the boundary between non-graph baselines and STGNNs. Right: mean VPT over valid rollouts.}
  \label{fig:overall_ranking_appendix}
\end{figure}

\paragraph{Aggregate performance summaries.}
Figure~\ref{fig:overall_ranking_appendix} shows the overall fractional win
rates and mean VPT across all twelve ranked models (Oracle GCN excluded as diagnostic control). Consistent with the main findings, D2STGNN and GWNet are close in overall wins, but this summary masks strong regime dependence. Mean VPT (right panel) provides a complementary view restricted to valid rollouts, emphasizing that overall rankings conflate conditional accuracy and robustness.

\begin{figure}[h]
  \centering
  \includegraphics[width=0.97\textwidth]{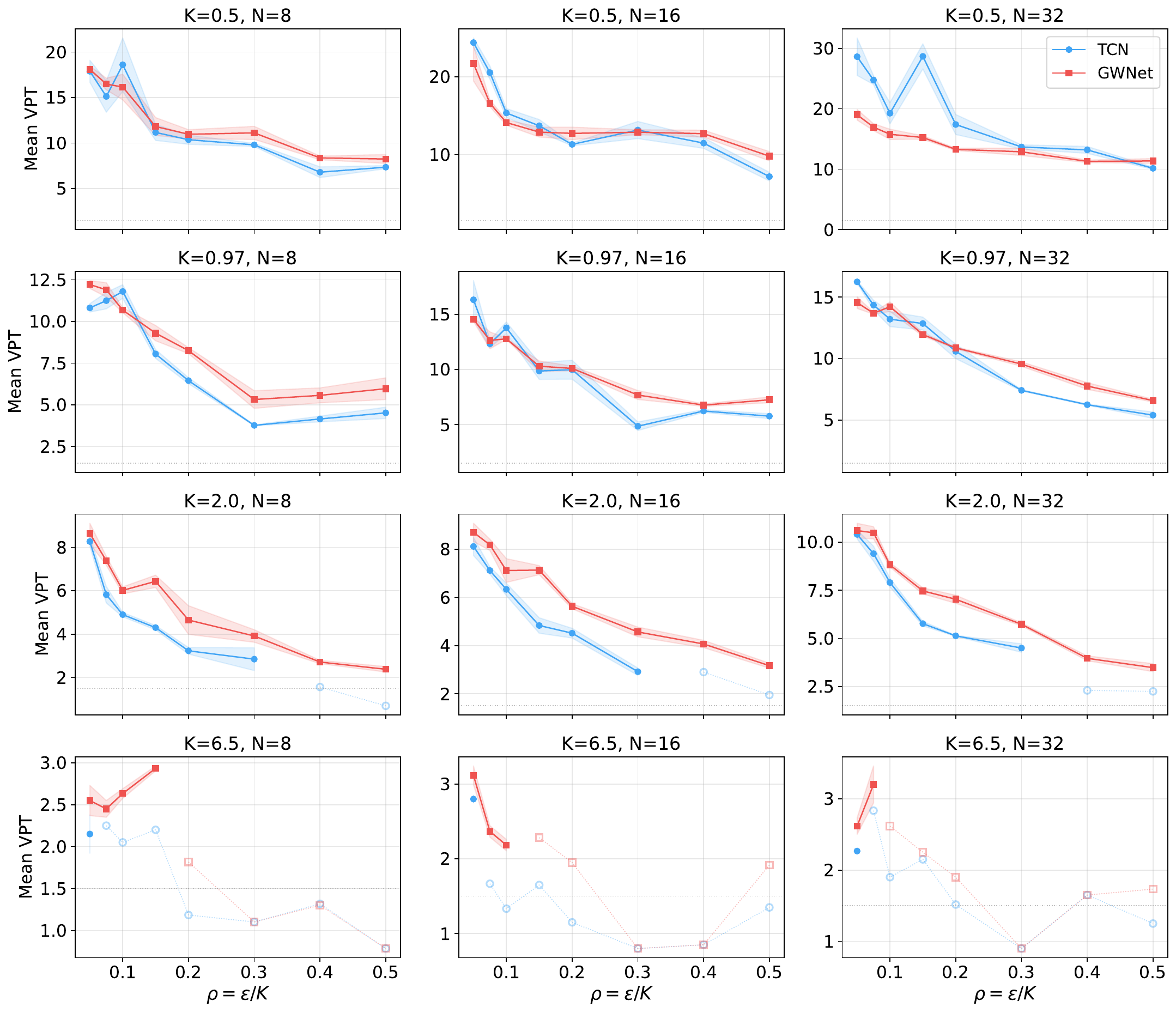}
  \caption{TCN vs.\ GWNet across $(K, N)$ regimes. At $K=0.5$ and $N=32$, local modeling dominates; as chaos increases, the crossover shifts toward smaller $\rho$ and the diffusion model becomes consistently reliable.}
  \label{fig:size_crossover_submission}
\end{figure}

Figure~\ref{fig:size_crossover_submission} further resolves the crossover across $(K, N)$ regimes. At low chaos ($K{=}0.5$), local models remain competitive, and at $N{=}32$ the matched TCN to GWNet comparison reverses across all $\rho$. As chaos increases, the crossover shifts downward in $\rho$ and diffusion based propagation becomes consistently beneficial across system sizes.

\begin{figure}[h]
  \centering
  \includegraphics[width=0.97\textwidth]{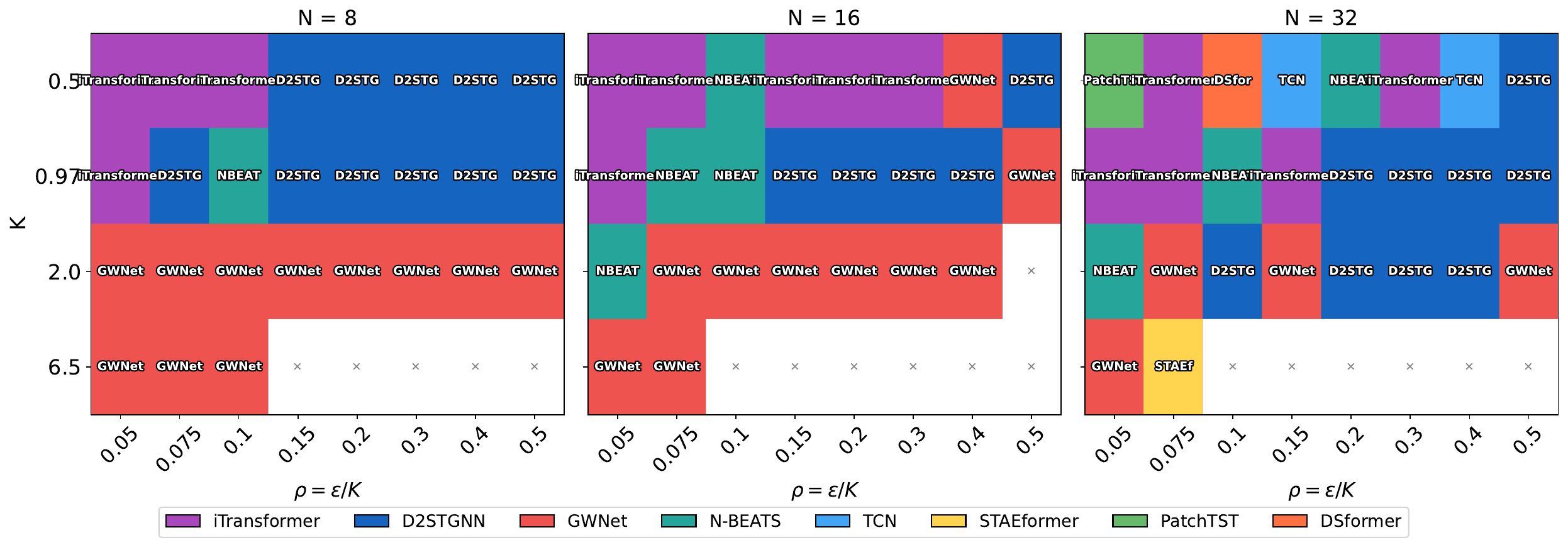}
  \caption{Per-instance winner map across the $(K, \rho, N)$ parameter grid (valid rollouts only). Each cell shows the model with the highest mean VPT over three seeds; ties are broken by VPT margin. The regime boundary where spatial models (GWNet, D2STGNN) displace local models (TCN, N-BEATS) shifts to lower $\rho$ as $K$ increases, consistent with the coupling-chaos crossover in Figure~\ref{fig:vpt_rho_crossover}.}
  \label{fig:winner_map}
\end{figure}

Figure~\ref{fig:winner_map} provides a complementary view of the coupling-chaos crossover by mapping the per-instance winning model across the full $(K, \rho, N)$ benchmark design space. Non-graph baselines (iTransformer, N-BEATS, TCN) dominate the low-$K$, low-$\rho$ corner, while graph-based STGNNs (GWNet, D2STGNN) become dominant as $K$ and $\rho$ increase. The boundary between these regions shifts toward lower $\rho$ as $K$ increases, consistent with the coupling-chaos crossover threshold reported in Table~\ref{tab:rho_c}.

\paragraph{Per-regime mean VPT.}
Mean VPT over valid rollouts mirrors the crossover while highlighting a
different aggregation. At $K{=}0.5$, ITrans leads ($16.93 \pm 7.02$), followed by N-BEATS ($16.27$), PatchTST ($15.75$), TCN ($15.41$), and D2STGNN ($13.22$). At $K{=}0.97$, D2STGNN leads ($10.20 \pm 2.44$), closely followed by GWNet ($10.02$), ITrans ($9.91$), and N-BEATS ($9.58$). At $K{=}2.0$, GWNet leads ($6.18 \pm 2.38$), followed by ITrans ($6.11$), TCN ($5.91$), and N-BEATS ($5.88$). At $K{=}6.5$, only low-$\rho$ system instances survive; GWNet leads ($2.67 \pm 0.34$), with STAEformer ($2.52 \pm 0.46$) and TCN ($2.41 \pm 0.35$) trailing on fewer instances. This progression reinforces that spatial mechanisms become more valuable as
coupling competes with increasing local chaos.

\paragraph{D2STGNN: mechanism fidelity vs.\ optimization fragility.}
D2STGNN explicitly decomposes diffusion and inherent dynamics, closely mirroring the governing system via a GRU-based recurrent evolution. This structural alignment yields strong conditional accuracy where the model produces valid rollouts, but also introduces sensitivity under autoregressive rollout. Empirically, performance peaks at moderate chaos: at $K{=}0.97$, D2STGNN achieves $\overline{\mathrm{VPT}} = 10.20 \pm 2.44$ and wins 14/24 scored system instances, while at $K{=}0.5$ it remains competitive ($13.22 \pm 2.08$). However, robustness degrades sharply as chaos increases: valid rollouts drop to 14/24 at $K{=}2.0$ and to 0/24 at $K{=}6.5$. GWNet succeeds on 19 system instances where D2STGNN fails, with no reverse cases.

Where both models produce valid rollouts at $K{=}2.0$, D2STGNN loses 4--10 and exhibits substantially higher seed variance ($\sigma$ up to 7.07 vs.\ 0.12--0.32 for GWNet), indicating instability rather than systematic accuracy differences. Across all 62 such instances, D2STGNN and GWNet tie 31--31 (Wilcoxon $p = 0.15$, n.s.), but the paired valid-rollout table (62 both, 0 D2 only, 19 GW only, 15 neither) yields a strong robustness gap (McNemar $p = 3.8 \times 10^{-6}$).

Overall, the results suggest that while mechanism level alignment improves performance when optimization succeeds, the gated recurrent decomposition also amplifies sensitivity under chaotic rollout, leading to reduced robustness relative to the implicitly parameterized diffusion model.

\paragraph{Failure-mode taxonomy under increasing chaos.}

Across model families, we observe three distinct breakdown mechanisms as chaos increases (Table~\ref{tab:failure_taxonomy}).

\begin{table}[h]
\centering
\caption{Failure-mode taxonomy under increasing chaos. Each model exhibits a distinct breakdown mechanism: representation distortion (DSformer), optimization instability (AGCRN), and spatial structure mismatch (STAEformer). Valid-rollout trend shows counts at $K{=}0.97/2.0/6.5$ (all three models achieve 24/24 valid rollouts at $K{=}0.5$); each regime has 24 system instances.}
\label{tab:failure_taxonomy}
\small
\begin{tabular}{lccp{0.30\textwidth}}
\toprule
Model & Failure class & Valid-rollout trend & Key evidence \\
\midrule
DSformer & Representation-level & 21 $\rightarrow$ 14 $\rightarrow$ 0 & RevIN removes scale; collapse at $K{=}6.5$ \\
AGCRN & Optimization-level & 24 $\rightarrow$ 18 $\rightarrow$ 0 & recurrent instability under rollout drift \\
STAEformer & Spatial structure mismatch & 24 $\rightarrow$ 22 $\rightarrow$ 7 & high valid rollout rate but 1.5/78 wins \\
\bottomrule
\end{tabular}
\end{table}

DSformer fails due to representation instability induced by Reversible Instance Normalization (RevIN), which standardizes each trajectory independently. While effective at low and moderate chaos (21/24 and 14/24 valid rollouts at $K{=}0.97$ and $K{=}2.0$), performance collapses at $K{=}6.5$ (0/24). This behavior is consistent with RevIN removing amplitude information required to locate trajectories on the chaotic attractor, an effect that becomes increasingly destructive as per-window variance grows with coupling strength.

AGCRN exhibits optimization and capacity failure: although it completes all runs, valid rollouts degrade monotonically (24/24 $\rightarrow$ 18/24 $\rightarrow$ 0/24) and vanish at high chaos. A controlled ablation under moderate chaos shows limited sensitivity to increased model capacity, suggesting that the failure is not purely due to under-parameterization but arises from instability in recurrent optimization under chaotic rollout dynamics. We also note that its sequential implementation limits parallelization relative to convolutional diffusion models such as GWNet.

STAEformer remains trainable across regimes but exhibits a mechanism mismatch under chaotic rollout, achieving only 1.5 wins out of 78 matched comparisons despite high robustness. This indicates that state dependent spatial attention, which recomputes propagation weights from evolving hidden states, is poorly aligned with the fixed coupling structure of the system, particularly under rapidly diverging trajectories. Taken together, these failure modes correspond respectively to representation-, optimization-, and spatial-structure mismatches in dynamical forecasting systems.

Across these cases, failures operate at different levels of the forecasting system: representation (DSformer), optimization (AGCRN), and propagation mechanism (STAEformer). Despite different origins, all exhibit increasing sensitivity to dynamical amplification as chaos strengthens, culminating in regime-specific collapse or degradation.

These failure modes correspond to distinct layers of sensitivity in forecasting systems under chaotic amplification, where error growth manifests as representation loss, optimization instability, or propagation instability depending on the model class.

\paragraph{Matched comparison: cross-variate attention (PatchTST vs.\ ITrans).}
Across the 66 instances where both models produce valid rollouts, ITrans wins 52, PatchTST wins 13, and 1 is a tie (Wilcoxon $p = 1.6 \times 10^{-7}$). The advantage is strongest at $K{=}0.5$, where relatively smooth dynamics preserve stable cross-variate correlations and implicit cross-variate mixing is sufficient. However, this advantage does not survive the full crossover: as chaos intensifies, ITrans is eventually overtaken by diffusion based models.

\paragraph{Matched comparison: diffusion convolution (TCN vs.\ GWNet).}
With similar temporal backbones, this comparison isolates the effect of diffusion convolution. At $K{=}0.5$, results are mixed. At $K{=}0.97$, GWNet wins 18/24 matched system instances; at $K{=}2.0$, it wins all 18/18 matched instances with valid rollouts from both models. The conditional accuracy gain (Wilcoxon $p = 3.3 \times 10^{-2}$, 69 pairs) is matched by a robustness advantage: GWNet produces valid rollouts on 12 instances where TCN fails, with no reverse cases (McNemar $p = 4.9 \times 10^{-4}$). The benefit of diffusion convolution grows monotonically with chaos and coupling severity (Figure~\ref{fig:tcn_gwnet_submission}).

\paragraph{Graph recovery: extended analysis.}
We evaluate whether learned adjacency structures recover the true ring topology underlying the CSM. Across 81 system instances with valid rollouts, GWNet achieves modest structural fidelity ($F_1 = 0.108$ and AUROC $= 0.657$), while AGCRN performs near chance ($F_1 = 0.019$, AUROC $= 0.546$). Despite this weak alignment with ground truth nearest neighbor coupling, GWNet consistently yields the strongest forecasting performance in the suite.

Structural alignment varies systematically with regime. It is highest in the near-integrable regime ($K{=}0.5$, AUROC $= 0.74$), where quasi-periodic dynamics preserve spatial correlations, and decreases near the onset of global chaos ($K{=}0.97$, AUROC $= 0.60$). In the strongest chaotic regime ($K{=}6.5$), the learned adjacency becomes almost uniform (weight ratio $1.1\times$ vs.\ $2.1\times$ at $K{=}0.5$). This rapid mixing effectively removes any persistent spatial structure within a few Lyapunov times. 

Despite using the exact ring topology, Oracle GCN does not outperform GWNet in forecasting, confirming that structural fidelity is neither necessary nor sufficient for predictive performance. Overall, learned adjacency reflects functional connectivity: dynamically useful correlations for prediction rather than physical coupling.

\paragraph{Diffusion mechanics: why the physics favors fixed spatial information propagation.}
The coupling term in Eq.~\ref{eq:csm} is nearest neighbor and diffusion-like: $\mathcal{C}[q_n]_i = \sin(q_n^{(i+1)} - q_n^{(i)}) + \sin(q_n^{(i-1)} - q_n^{(i)})$. For small phase differences this reduces to a discrete Laplacian on the ring; more generally it acts as local smoothing of phase differences.

This local interaction structure is consistent with diffusion based message passing, which aggregates information over neighboring nodes using a fixed spatial structure learned during training. In contrast, spatial attention mechanisms are more flexible but do not explicitly encode locality or diffusion structure.
 
\begin{figure}[ht]
  \centering
  \includegraphics[width=0.85\textwidth]{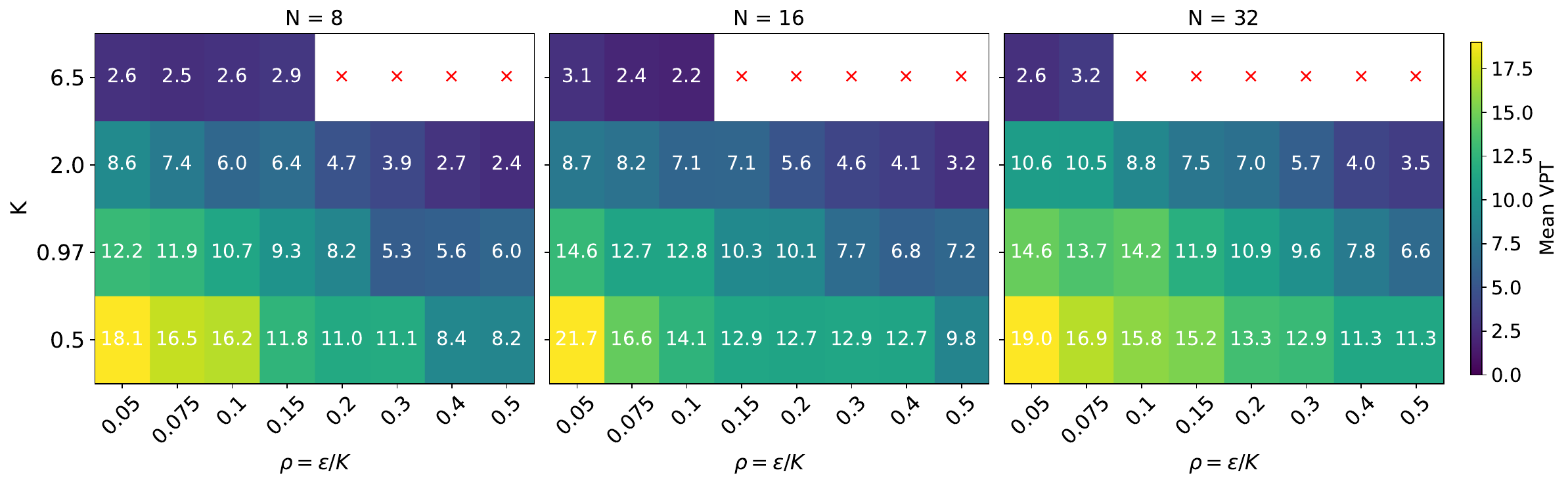}
  \caption{Autoregressive VPT for GWNet across the $(K, \rho)$ parameter space at three system sizes ($N \in \{8, 16, 32\}$). Yellow indicates high VPT (long valid prediction), purple indicates low VPT. The monotonic decay with increasing $K$ and $\rho$ reflects the coupling-chaos crossover. The value of spatial structure increases with coupling, while chaos reduces overall predictability across all regimes.}
  \label{fig:vpt_heatmap_gwnet}
\end{figure}

Under autoregressive rollout, this distinction becomes critical. GWNet's fixed adjacency preserves stable spatial information propagation, ensuring that prediction errors do not propagate through changing spatial weights. STAEformer recomputes spatial attention from the current evolving hidden states at every step, creating a feedback loop in which prediction errors perturb hidden states, which in turn perturb propagation weights, amplifying drift. This effect becomes increasingly severe in chaotic regimes where state perturbations grow rapidly over short Lyapunov times. Figure~\ref{fig:vpt_heatmap_gwnet} summarizes this decay across the full parameter space.

\section{Computational Resources}
\label{app:compute}

The main sweep was conducted on NVIDIA A100-80GB PCIe GPUs (institutional HPC
cluster, 32 CPU cores, 62\,GB RAM per node).
Estimated total compute: ${\sim}4{,}500$ A100 GPU-hours
(3 seeds $\times$ 13 models $\times$ 96 system instances; per-run training
times range from ${\sim}10$\,min for DLinear to ${\sim}5$\,h for AGCRN and
STAEformer).

\section{Broader Impact}
\label{app:impact}

This benchmark is based on a well studied dynamical system and does not involve human subjects, sensitive data, or direct use concerns.

The primary contribution is methodological: by identifying when and why spatial structure improves forecasting under chaotic dynamics, the framework can guide more principled model selection and reduce unnecessary computational cost.

A potential risk is misinterpretation of results beyond their intended scope. While the benchmark reveals mechanism dependent behavior in controlled systems, real world datasets may exhibit different or unknown generating dynamics. We therefore view this framework as complementary to application-specific evaluation rather than a substitute for it. More broadly, the framework may contribute to more reliable deployment of forecasting models in scientific and engineering settings where robustness under rollout is critical.

\section{Reproducibility}
\label{app:release}

The \texttt{chaosnetbench} Python package (MIT license) is publicly available at \url{https://github.com/htmoges/ChaosNetBench}, including data generation, training, evaluation, figure reproduction scripts, and Croissant~\citep{akhtar2024croissant} metadata records.

The public release documents access to the full dataset (27.3\,GB, \texttt{chaosnetbench\_cml.h5}), precomputed benchmark results, and reproduction workflows from quick tests to the full benchmark sweep. Three reproduction tiers are documented: \texttt{test} ($\sim$10\,min, one model, one system instance), \texttt{medium} (one model, one $K$ slice, $\sim$2\, GPU-hours), and \texttt{full} (13 models $\times$ 96 configs $\times$ 3 seeds, $\sim$4{,}500\,GPU-hours on NVIDIA A100-80\,GB).

\section{Glossary of Terms}
\label{app:glossary}

The following terms are used consistently throughout this paper.

\begin{description}[leftmargin=2em,labelindent=0em,itemsep=4pt]
  \item[\textbf{System instance.}] A configuration defined by $(K, \rho, N)$. Each system instance comprises 100 trajectories generated from distinct ICs.

\item[\textbf{Benchmark design space.}] The grid of controlled parameters $(K, \rho, N)$ used to generate system instances.

\item[\textbf{Controlled factors.}] The benchmark variables set directly by design: $K$ (local chaos), $\rho = \varepsilon/K$ (coupling-to-chaos ratio), and $N$ (system size).

\item[\textbf{Valid rollout.}] A forecast that remains informative rather than collapsing to a mean-state prediction, screened by a test MSE threshold (see Table~\ref{tab:setup}).

\item[\textbf{Conditional accuracy.}] Predictive performance, measured using VPT, evaluated only on system instances where both models under comparison produce valid rollouts.

\item[\textbf{Robustness.}] The fraction of system instances for which a model produces a valid rollout.

\item[\textbf{Averaged accuracy.}] A single summary error (e.g., MSE or RMSE) computed across all system instances, without screening for valid rollouts.

\item[\textbf{Chaos indicators.}] Post hoc diagnostics computed from trajectories to characterize dynamical difficulty, independent of model predictions. Includes $\lambda_{\max}$ (maximal Lyapunov exponent) and SALI (Smaller Alignment Index).

\item[\textbf{Trajectory divergence.}] Exponential separation of nearby trajectories due to chaotic dynamics, quantified by $\lambda_{\max}$.

\item[\textbf{Forecast error growth.}] Increase in prediction error as rollout is extended, amplified by trajectory divergence.
\end{description}

\end{document}